\documentclass[12pt,authoryear,preprint]{elsarticle}

\DeclareGraphicsExtensions{.pdf,.jpeg,.png}
\usepackage[caption = false,font = footnotesize]{subfig}
\usepackage{cite}
\usepackage{url}
\usepackage{rotating}

\usepackage[cmex10]{amsmath}
\usepackage{amsfonts}
\usepackage{amssymb}
\usepackage{amsthm}
\usepackage{bm}
\usepackage{empheq}
\usepackage{booktabs}

\usepackage{algorithmic}
\usepackage{setspace}
\usepackage{array}

\usepackage{url}
\usepackage[table]{xcolor}
\usepackage{multirow}
\usepackage{booktabs}
\usepackage{flushend}
\usepackage[colorlinks=true]{hyperref}
\usepackage{float}
\usepackage[flushleft]{threeparttable} 

\newsubfloat{figure}

\newcommand{\vect}[1]{\mathbf{#1}}

\DeclareMathSymbol{\R}{\mathalpha}{AMSb}{"52}

\newcommand{\norm}[2][2]{\left\lVert #2 \right\rVert_{#1}}

\newtheorem{theorem}{Theorem}

\journal{Neural Networks}

\begin{document}

\begin{frontmatter}

\title{Kafnets: kernel-based non-parametric activation functions for neural networks}

\author[sapienza]{Simone Scardapane\corref{cor1}}
\ead{simone.scardapane@uniroma1.it}
\cortext[cor1]{Corresponding author. Phone: +39 06 44585495, Fax: +39 06 4873300.}

\author[cantabria]{Steven Van Vaerenbergh}
\ead{steven.vanvaerenbergh@unican.es}

\author[sapienza2]{Simone Totaro}

\author[sapienza]{Aurelio Uncini}
\ead{aurelio.uncini@uniroma1.it}

\address[sapienza]{Department of Information Engineering, Electronics and Telecommunications (DIET), Sapienza University of Rome, Via Eudossiana 18, 00184 Rome, Italy}

\address[sapienza2]{Department of Statistical Sciences, Sapienza University of Rome, Piazzale Aldo Moro 5, 00185 Rome, Italy.}

\address[cantabria]{Department of Communications Engineering, University of Cantabria, Av. los Castros s/n, 39005 Santander, Cantabria, Spain.}

\begin{abstract}
Neural networks are generally built by interleaving (adaptable) linear layers with (fixed) nonlinear activation functions. To increase their flexibility, several authors have proposed methods for adapting the activation functions themselves, endowing them with varying degrees of flexibility. None of these approaches, however, have gained wide acceptance in practice, and research in this topic remains open. In this paper, we introduce a novel family of flexible activation functions that are based on an inexpensive kernel expansion at every neuron. Leveraging over several properties of kernel-based models, we propose multiple variations for designing and initializing these kernel activation functions (KAFs), including a multidimensional scheme allowing to nonlinearly combine information from different paths in the network. The resulting KAFs can approximate any mapping defined over a subset of the real line, either convex or nonconvex. Furthermore, they are smooth over their entire domain, linear in their parameters, and they can be regularized using any known scheme, including the use of $\ell_1$ penalties to enforce sparseness. To the best of our knowledge, no other known model satisfies all these properties simultaneously. In addition, we provide a relatively complete overview on alternative techniques for adapting the activation functions, which is currently lacking in the literature. A large set of experiments validates our proposal.
\end{abstract}


\end{frontmatter}

\section{Introduction}
\label{sec:introduction}

Neural networks (NNs) are powerful approximators, which are built by interleaving linear layers with nonlinear mappings (generally called \emph{activation functions}). The latter step is usually implemented using an element-wise, (sub-)differentiable, and fixed nonlinear function at every neuron. In particular, the current consensus has shifted from the use of contractive mappings (e.g., sigmoids) to the use of piecewise-linear functions (e.g., rectified linear units, ReLUs \citep{glorot2010understanding}), allowing a more efficient flow of the backpropagated error \citep{goodfellow2016deep}. This relatively inflexible architecture might help explaining the extreme redundancy found in the trained parameters of modern NNs \citep{denil2013predicting}.

Designing ways to adapt the activation functions themselves, however, faces several challenges. On one hand, we can parameterize a known activation function with a small number of trainable parameters, describing for example the slope of a particular linear segment \citep{he2015delving}. While immediate to implement, this only results in a small increase in flexibility and a marginal improvement in performance in the general case \citep{agostinelli2014learning}. On the other hand, a more interesting task is to devise a scheme allowing each activation function to model a large range of shapes, such as any smooth function defined over a subset of the real line. In this case, the inclusion of one (or more) hyper-parameters enables the user to trade off between greater flexibility and a larger number of parameters per-neuron. We refer to these schemes in general as \emph{non-parametric} activation functions, since the number of (adaptable) parameters can potentially grow without bound.

There are three main classes of non-parametric activation functions known in the literature: adaptive piecewise linear (APL) functions \citep{agostinelli2014learning}, maxout networks \citep{goodfellow2013maxout}, and spline activation functions \citep{guarnieri1999multilayer}. These are described more in depth in Section \ref{sec:nonparametric_afs}. In there, we also argue that none of these approaches is fully satisfactory, meaning that each of them loses one or more desirable properties, such as smoothness of the resulting functions in the APL and maxout cases (see Table \ref{tab:non_parametric_afs_comparison} in Section \ref{sec:proposed_kernel_afs} for a schematic comparison).

In this paper, we propose a fourth class of non-parametric activation functions, which are based on a kernel representation of the function. In particular, we define each activation function as a linear superposition of several kernel evaluations, where the dictionary of the expansion is fixed beforehand by sampling the real line. As we show later on, the resulting kernel activation functions (KAFs) have a number of desirable properties, including: (i) they can be computed cheaply using vector-matrix operations; (ii) they are linear with respect to the trainable parameters; (iii) they are smooth over their entire domain; (iv) using the Gaussian kernel, their parameters only have \textit{local} effects on the resulting shapes; (v) the parameters can be regularized using any classical approach, including the possibility of enforcing sparseness through the use of $\ell_1$ norms. To the best of our knowledge, none of the known methods possess all these properties simultaneously. We call a NN endowed with KAFs at every neuron a \textit{Kafnet}.

Importantly, framing our method as a kernel technique allows us to potentially leverage over a huge literature on kernel methods, either in statistics, machine learning \citep{hofmann2008kernel}, and signal processing \citep{liu2011kernel}. Here, we preliminarly demonstrate this by discussing several heuristics to choose the kernel hyper-parameter, along with techniques for initializing the trainable parameters. However, much more can be applied in this context, as we discuss in depth in the conclusive section. We also propose a bi-dimensional variant of our KAF, allowing the information from multiple linear projections to be nonlinearly combined in an adaptive fashion. 

In addition, we contend that one reason for the rareness of flexible activation functions in practice can be found in the lack of a cohesive (introductory) treatment on the topic. To this end, a further aim of this paper is to provide a relatively comprehensive overview on the selection of a proper activation function. In particular, we divide the discussion on the state-of-the-art in three separate sections. Section \ref{sec:fixed_af} introduces the most common (fixed) activation functions used in NNs, from the classical sigmoid function up to the recently developed self-normalizing unit \citep{klambauer2017self} and Swish function \citep{ramachandran2017swish}. Then, we describe in Section \ref{sec:parametric_af} how most of these functions can be efficiently parameterized by one or more adaptive scalar values in order to enhance their flexibility. Finally, we introduce the three existing models for designing non-parametric activation functions in Section \ref{sec:nonparametric_afs}. For each of them, we briefly discuss relative strengths and drawbacks, which serve as a motivation for introducing the model subsequently.

The rest of the paper is composed of four additional sections. Section \ref{sec:proposed_kernel_afs} describes the proposed KAFs, together with several practical implementation guidelines regarding the selection of the dictionary and the choice of a proper initialization for the weights. For completeness, Section \ref{sec:related_work} briefly describes additional strategies to improve the activation functions, going beyond the addition of trainable parameters in the model. A large set of experiments is described in Section \ref{sec:experiments}, and, finally, the main conclusions and a set of future lines of research are given in Section \ref{sec:conclusions}.

\subsubsection*{Notation}
We denote vectors using boldface lowercase letters, e.g., $\vect{a}$; matrices are denoted by boldface uppercase letters, e.g., $\vect{A}$. All vectors are assumed to be column vectors. The operator $\norm[p]{\cdot}$ is the standard $\ell_p$ norm on an Euclidean space. For $p=2$, it coincides with the Euclidean norm, while for $p=1$ we obtain the Manhattan (or taxicab) norm defined for a generic vector $\vect{v} \in \R^B$ as $\norm[1]{\vect{v}} = \sum_{k=1}^B |v_k|$. Additional notations are introduced along the paper when required.

\section{Preliminaries}
\label{sec:preliminaries}

We consider training a standard feedforward NN, whose $l$-th layer is described by the following equation:
\begin{equation}
\vect{h}_l = g_l\left( \vect{W}_l\vect{h}_{l-1} + \vect{b}_l \right) \,,
\label{eq:nn_layer}
\end{equation}
where $\vect{h}_{l-1} \in \R^{N_{l-1}}$ is the $N_{l-1}$-dimensional input to the layer, $\vect{W}_l \in \R^{N_{l-1} \times N_l}$ and $\vect{b}_l \in \R^{N_l}$ are adaptable weight matrices, and $g_l(\cdot)$ is a nonlinear function, called \emph{activation function}, which is applied element-wise. In a NN with $L$ layers, $\vect{x} = \vect{h}_0$ denotes the input to the network, while $\hat{\vect{y}} = \vect{h}_L$ denotes the final output.

For training the network, we are provided with a set of $I$ input/output pairs $\mathcal{S} = \left\{ \vect{x}_i, \vect{y}_i \right\}_{i=1}^I$, and we minimize a regularized cost function given by:
\begin{equation}
J(\vect{w}) = \sum_{i=1}^I l(\vect{y}_i, \hat{\vect{y}}_i) + C \cdot r(\vect{w}) \,,
\end{equation}
where $\vect{w} \in \R^Q$ collects all the trainable parameters of the network, $l(\cdot, \cdot)$ is a loss function (e.g., the squared loss), $r(\cdot)$ is used to regularize the weights using, e.g., $\ell_2$ or $\ell_1$ penalties, and the regularization factor $C > 0$ balances the two terms. 


In the following, we review common choices for the selection of $g_l$, before describing methods to adapt them based on the training data. For readability, we will drop the subscript $l$, and we use the letter $s$ to denote a single input to the function, which we call an \emph{activation}. Note that, in most cases, the activation function $g_L$ for the last layer cannot be chosen freely, as it depends on the task and a proper scaling of the output. In particular, it is common to select $g(s) = s$ for regression problems, and a sigmoid function for binary problems with $y_i \in \left\{0, 1\right\}$:
\begin{equation}
g(s) = \delta(s) = \frac{1}{1+\exp\left\{-s\right\}}\,.
\label{eq:sigmoid}
\end{equation}
For multi-class problems with dummy encodings on the output, the softmax function generalizes the sigmoid and it ensures valid probability distributions in the output \citep{bishop2006pattern}.

\section{Fixed activation functions}
\label{sec:fixed_af}

We briefly review some common (fixed) activation functions for neural networks, that are the basis for the parametric ones in the next section. Before the current wave of deep learning, most activation functions used in NNs were of a `squashing' type, i.e., they were monotonically non-decreasing functions satisfying:
\begin{equation}
\lim_{s \rightarrow -\infty} g(s) = c, \,\, \lim_{s \rightarrow \infty} = 1 \,,
\end{equation}
where $c$ can be either $0$ or $-1$, depending on convention. Apart from the sigmoid in \eqref{eq:sigmoid}, another common choice is the hyperbolic tangent, defined as:
\begin{equation}
g(s) = \text{tanh}(s) = \frac{\exp\left\{s\right\} - \exp\left\{-s\right\}}{\exp\left\{s\right\} + \exp\left\{-s\right\}} \,.
\label{eq:tanh}
\end{equation}

%


\noindent \citet{cybenko1989approximation} proved the universal approximation property for this class of functions, and his results were later extended to a larger class of functions in \citet{hornik1989multilayer}. In practice, squashing functions were found of limited use in deep networks (where $L$ is large), being prone to the problem of vanishing and exploding gradients, due to their bounded derivatives \citep{hochreiter2001gradient}. A breakthrough in modern deep learning came from the introduction of the rectifier linear unit (ReLU) function, defined as:


\begin{equation}
g(s) = \max\left\{0, s\right\} \,.
\end{equation}
Despite being unbounded and introducing a point of non-differentiability, the ReLU has proven to be extremely effective for deep networks \citep{glorot2010understanding,maas2013rectifier}. The ReLU has two main advantages. First, its gradient is either $0$ or $1$,\footnote{For $s=0$, the function is not differentiable, but any value in $\left[0,1\right]$ is a valid subgradient. Most implementations of ReLU use $0$ as the default choice in this case.} making back-propagation particularly efficient. Secondly, its activations are sparse, which is beneficial from several points of views. A smoothed version of the ReLU, called softplus, is also introduced in \citet{glorot2011deep}:


\begin{equation}
g(s) = \log\left\{ 1 + \exp\left\{ s \right\} \right\} \,.
\end{equation}
Despite its lack of smoothness, ReLU functions are almost always preferred to the softplus in practice. One obvious problem of ReLUs is that, for a wrong initialization or an unfortunate weight update, its activation can get stuck in $0$, irrespective of the input. This is referred to as the `dying ReLU' condition. To circumvent this problem, \citet{maas2013rectifier} introduced the leaky ReLU function, defined as:
\begin{equation}
g(s) = 
	\begin{cases}
	s & \text{ if } s \ge 0 \\
	\alpha s & \text{ otherwise }
	\end{cases} \,,
\label{eq:leaky_relu}
\end{equation}
where the user-defined parameter $\alpha > 0$ is generally set to a small value, such as $0.01$. While the resulting pattern of activations is not exactly sparse anymore, the parameters cannot get stuck in a poor region. \eqref{eq:leaky_relu} can also be written more compactly as $ g(s) = \max\left\{ 0, s \right\} + \alpha \min\left\{ 0, s \right\}$.

Another problem of activation functions having only non-negative output values is that their mean value is always positive by definition. Motivated by an analogy with the natural gradient, \citet{clevert2016fast} introduced the exponential linear unit (ELU) to renormalize the pattern of activations:


\begin{equation}
g(s) = \text{ELU}(s) 
	\begin{cases}
	s & \text{ if } s \ge 0 \\
	\alpha \left( \exp\left\{ s \right\} - 1 \right) & \text{ otherwise }
	\end{cases} \,,
\label{eq:elu}
\end{equation}
where in this case $\alpha$ is generally chosen as $1$. The ELU modifies the negative part of the ReLU with a function saturating at a user-defined value $\alpha$. It is computationally efficient, being smooth (differently from the ReLU and the leaky ReLU), and with a gradient which is either $1$ or $g(s)+\alpha$ for negative values of $s$.

The recently introduced scaled ELU (SELU) generalizes the ELU to have further control over the range of activations \citep{klambauer2017self}:
\begin{equation}
g(s) = \text{SELU}(s) = \lambda \cdot \text{ELU}(s) \,,
\end{equation}
where $\lambda > 1$ is a second user-defined parameter. Particularly, it is shown in \citet{klambauer2017self} that for $\lambda \approx 1.6733$ and $\alpha \approx 1.0507$, the successive application of \eqref{eq:nn_layer} converges towards a fixed distribution with zero mean and unit variance, leading to a self-normalizing network behavior.

Finally, Swish \citep{ramachandran2017swish} is a recently proposed activation somewhat inspired by the gating steps in a standard LSTM recurrent cell:
\begin{equation}
g(s) = s \cdot \delta(s) \,,
\end{equation}
where $\delta(s)$ is the sigmoid in \eqref{eq:sigmoid}.

\section{Parametric adaptable activation functions}
\label{sec:parametric_af}

An immediate way to increase the flexibility of a NN is to parameterize one of the previously introduced activation functions with a \textit{fixed} (small) number of adaptable parameters, such that each neuron can adapt its activation function to a different shape. As long as the function remains differentiable with respect to these new parameters, it is possible to adapt them with any numerical optimization algorithm together with the linear weights and biases of the layer. Due to their fixed number of parameters and limited flexibility, we call these \textit{parametric} activation functions.

Historically, one of the first proposals in this sense was the generalized hyperbolic tangent \citep{chen1996feedforward}, a tanh function parameterized by two additional positive scalar values $a$ and $b$:
\begin{equation}
g(s) = \frac{a \left( 1 - \exp\left\{- bs\right\}\right)}{1 + \exp\left\{-bs\right\}} \,.
\label{eq:generalized_tanh}
\end{equation}
Note that the parameters $a,b$ are initialized randomly and are adapted independently for every neuron. Specifically, $a$ determines the range of the output (which is called the amplitude of the function), while $b$ controls the slope of the curve. \citet{trentin2001networks} provides empirical evidence that learning the amplitude for each neuron is beneficial (either in terms of generalization error, or speed of convergence) with respect to having unit amplitude for all activation functions. Similar results were also obtained for recurrent networks \citep{goh2003recurrent}.

More recently, \citet{he2015delving} consider a parametric version of the leaky ReLU in \eqref{eq:leaky_relu}, where the coefficient $\alpha$ is initialized at $\alpha = 0.25$ everywhere and then adapted for every neuron. The resulting activation function is called parametric ReLU (PReLU), and it has a very simple derivative with respect to the new parameter:
\begin{equation}
\frac{dg(s)}{d\alpha} = 
	\begin{cases}
	0 & \text{ if } s \ge 0 \\
	s & \text{ otherwise }
	\end{cases} \,.
\end{equation}
For a layer with $N$ hidden neurons, this introduces only $N$ additional parameters, compared to $2N$ parameters for the generalized tanh. Importantly, in the case of $\ell_p$ regularization, the user has to be careful not to regularize the $\alpha$ parameters, which would bias the optimization process towards classical ReLU / leaky ReLU activation functions.

Similarly, \citet{trottier2016parametric} propose a modification of the ELU function in \eqref{eq:elu} with an additional scalar parameter $\beta$, called parametric ELU (PELU):
\begin{equation}
g(s) = 
	\begin{cases}
	\displaystyle\frac{\alpha}{\beta}s & \text{ if } s \ge 0 \\
	\alpha \left( \exp\left\{ \displaystyle\frac{s}{\beta} \right\} - 1 \right) & \text{ otherwise }
	\end{cases} \,,
\label{eq:pelu}
\end{equation}
where both $\alpha$ and $\beta$ are initialized randomly and adapted during the training process. Based on the analysis in \citet{jin2016deep}, there always exists a setting for the linear weights and $\alpha, \beta$ which avoids the vanishing gradient problem. Differently from the PReLU, however, the two parameters should be regularized in order to avoid a degenerate behavior with respect to the linear weights, where extremely small linear weights are coupled with very large values for the parameters of the activation functions.

A more flexible proposal is the S-shaped ReLU (SReLU) \citep{jin2016deep}, which is parameterized by four scalar values $\left\{ t^r, a^r, t^l, a^l \right\} \in \R^4$:
\begin{equation}
g(s) = 
	\begin{cases}
	t^r + a^r \left( s - t^r \right) & \text{ if } s \ge t^r \\
	s								 & \text{ if } t^r > s > t^l \\
	t^l + a^l \left(s - t^l \right)  & \text{ otherwise }
	\end{cases} \,.
\label{eq:srelu}
\end{equation}
The SReLU is composed by three linear segments, the middle of which is the identity. Differently from the PReLU, however, the cut-off points between the three segments can also be adapted. Additionally, the function can have both convex and nonconvex shapes, depending on the orientation of the left and right segments, making it more flexible than previous proposals. Similar to the PReLU, the four parameters should not be regularized \citep{jin2016deep}.

Finally, a parametric version of the Swish function is the $\beta$-swish \citep{ramachandran2017swish}, which includes a tunable parameter $\beta$ inside the self-gate:
\begin{equation}
g(s) = s \cdot \delta(\beta s) \,.
\end{equation}

\section{Non-parametric activation functions}
\label{sec:nonparametric_afs}

Intuitively, parametric activation functions have limited flexibility, resulting in mixed performance gains on average. Differently from parametric approaches, non-parametric activation functions allow to model a larger class of shapes (in the best case, any continuous segment), at the price of a larger number of adaptable parameters. As stated in the introduction, these methods generally introduce a further global hyper-parameter allowing to balance the flexibility of the function, by varying the effective number of free parameters, which can potentially grow without bound. Additionally, the methods can be grouped depending on whether each parameter has a local or global effect on the overall function, the former being a desirable characteristic.

In this section, we describe three state-of-the-art approaches for implementing non-parametric activation functions: APL functions in Section \ref{sec:apl_af}, spline functions in Section \ref{sec:spline_af}, and maxout networks in Section \ref{sec:maxout_networks}.

\subsection{Adaptive piecewise linear methods}
\label{sec:apl_af}

An APL function, introduced in \citet{agostinelli2014learning}, generalizes the SReLU function in \eqref{eq:srelu} by summing multiple linear segments, where all slopes and cut-off points are learned under the constraint that the overall function is continuous:
\begin{equation}
g(s) = \max\left\{0, s\right\} + \sum_{i=1}^S a_i \max\left\{0, -s + b_i \right\} \,.
\label{eq:apl}
\end{equation}
$S$ is a hyper-parameter chosen by the user, while each APL is parameterized by $2S$ adaptable parameters $\left\{ a_i, b_i \right\}_{i=1}^S$. These parameters are randomly initialized for each neuron, and can be regularized with $\ell_2$ regularization, similarly to the PELU, in order to avoid the coupling of very small linear weights and very large $a_i$ coefficients for the APL units.

The APL unit cannot approximate any possible function. Its approximation properties are described in the following theorem.
\begin{theorem}[Theorem 1, \citep{agostinelli2014learning}]
The APL unit can approximate any continuous piecewise-linear function $h(s)$, for some choice of $S$ and $\left\{ a_i, b_i \right\}_{i=1}^S$, provided that $h(s)$ satisfies the following two conditions:
\begin{enumerate}
\item There exists $u \in \R$ such that $h(s) = s, \,\, \forall s \ge u$.
\item There exist two scalars, $v, t \in \R$ such that $\frac{dh(s)}{s} = t, \,\, \forall s < v$.
\end{enumerate}
\end{theorem}
\noindent The previous theorem implies that any piecewise-linear function can be approximated, provided that its behavior is linear for very large, or small, $s$. A possible drawback of the APL activation function is that it introduces $S+1$ points of non-differentiability for each neuron, which may damage the optimization algorithm. The next class of functions solves this problem, at the cost of a possibly larger number of parameters.

\subsection{Spline activation functions}
\label{sec:spline_af}

An immediate way to exploit polynomial interpolation in NNs it to build the activation function over powers of the activation $s$ \citep{piazza1992artificial}:
\begin{equation}
g(s) = \sum_{i=0}^P a_i s^i \,,
\end{equation}
where $P$ is a hyper-parameter and we adapt the $(P+1)$ coefficients $\left\{a_i\right\}_{i=0}^P$. Since a polynomial of degree $P$ can pass exactly through $P+1$ points, this polynomial activation function (PAF) can in theory approximate any smooth function. The drawback of this approach is that each parameter $a_i$ has a global influence on the overall shape, and the output of the function can easily grow too large or encounter numerical problems, particularly for large absolute values of $s$ and large $P$.

An improved way to use polynomial expansions is spline interpolation, giving rise to the spline activation function (SAF). The SAF was originally studied in \citet{vecci1998learning,guarnieri1999multilayer}, and later re-introduced in a more modern context in \citet{scardapane2016learning}, following previous advances in nonlinear filtering \citep{scarpiniti2013nonlinear}. In the sequel, we adopt the newer formulation.

\begin{figure}
\subfloat[CR matrix]{
\includegraphics[width=0.5\columnwidth,keepaspectratio]{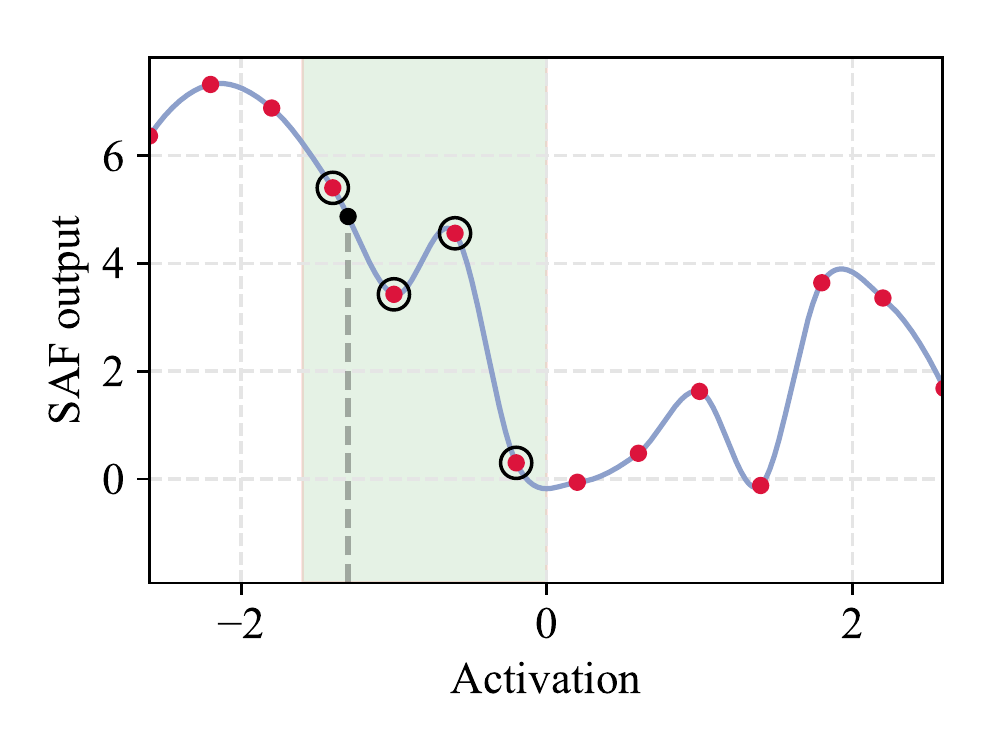}
\label{fig:saf_example_crbasis}
} \hfill
\subfloat[B-basis matrix]{
\includegraphics[width=0.5\columnwidth,keepaspectratio]{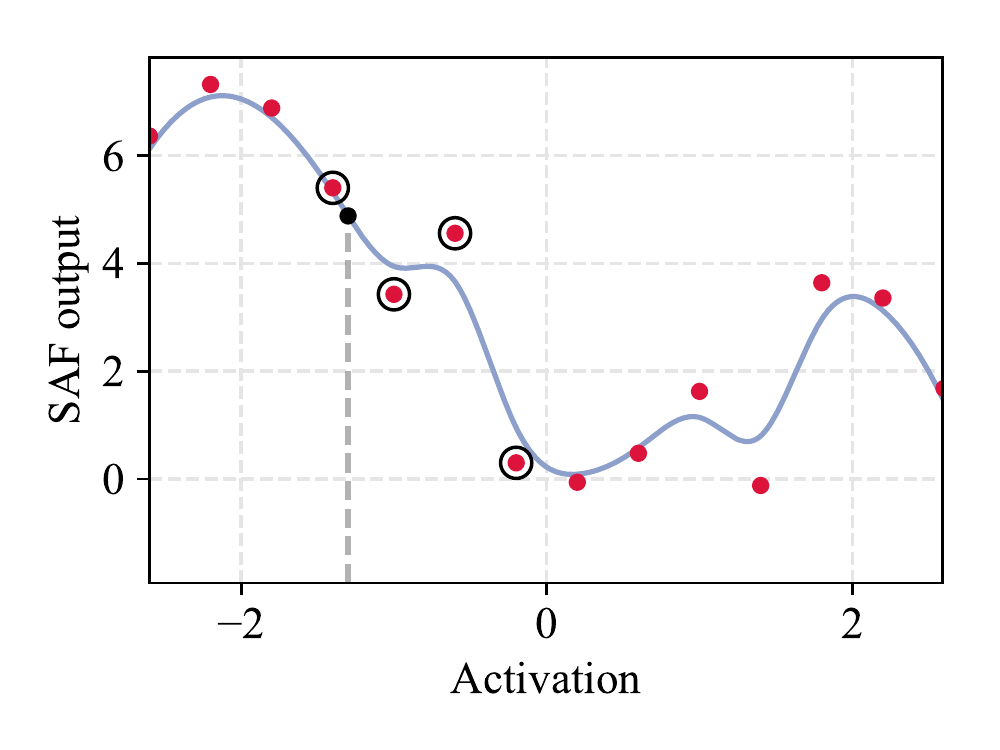}
\label{fig:saf_example_bbasis}
} \hfill
\caption{Example of output interpolation using a SAF neuron. Knots are shown with red markers, while the overall function is given in a light blue. (a) For a given activation, in black, only the control points in the green background are active. (b) We use the same control points as before, but we interpolate using the B-basis matrix \citep{scarpiniti2013nonlinear} instead of the CR matrix in \eqref{eq:catmulrom_basis_matrix}. The resulting curve is smoother, but it is not guaranteed to pass through all the control points.}
\label{fig:saf_example}
\end{figure}

A SAF is described by a vector of $T$ parameters, called \textit{knots}, corresponding to a sampling of its $y$-values over an equispaced grid of $T$ points over the $x$-axis, that are symmetrically chosen around the origin with sampling step $\Delta x$. For any other value of $s$, the output value of the SAF is computed with spline interpolation over the closest knot and its $P$ rightmost neighbors, where $P$ is generally chosen equal to $3$, giving rise to a cubic spline interpolation scheme. Specifically, denote by $k$ the index of the closest knot, and by $\vect{q}_k$ the vector comprising the corresponding knot and its $P$ neighbors. We call this vector the \textit{span}. We also define a new value
\begin{equation}
u = \frac{s}{\Delta x} - \left\lfloor \frac{s}{\Delta x} \right\rfloor \,,
\end{equation}
where $\Delta x$ is the user-defined sampling step. $u$ defines a normalized abscissa value between the $k$-th knot and the $(k+1)$-th one. The output of the SAF is then given by \citep{scarpiniti2013nonlinear}:
\begin{equation}
g(s) = \vect{u}^T \vect{B} \vect{q}_k \,,
\end{equation}
where the vector $\vect{u}$ collects powers of $u$ up to the order $P$:
\begin{equation}
\vect{u} = \left[ u^P, u^{P-1}, \ldots, u^1, 1 \right]^T \,,
\end{equation}
and $\vect{B}$ is the spline basis matrix, which defines the properties of the interpolation scheme (as shown later in Fig. \ref{fig:saf_example_bbasis}). For example, the popular Catmull-Rom basis for $P=3$ is given by:
\begin{equation}
\vect{B} = \frac{1}{2} 
\begin{bmatrix}
	-1 & 3 & -3 & 1 \\
	2 & -5 & 4 & -1 \\
	-1 & 0 & 1 & 0 \\
	0 & 2 & 0 & 0
\end{bmatrix} \,.
\label{eq:catmulrom_basis_matrix}
\end{equation}
The derivatives of the SAF can be computed in a similar way, both with respect to $s$ and with respect to $\vect{q}_k$, e.g., see \citet{scardapane2016learning}. A visual example of the SAF output is given in Fig. \ref{fig:saf_example}.

Each knot has only a limited local influence over the output, making their adaptation more stable. The resulting function is also smooth, and can in fact approximate any smooth function defined over a subset of the real line to a desired level of accuracy, provided that $\Delta x$ is chosen small enough. The drawback is that regularizing the resulting activation functions is harder to achieve, since $\ell_p$ regularization cannot be applied directly to the values of the knots. In \citet{guarnieri1999multilayer}, this was solved by choosing a large $\Delta x$, in turn severely limiting the flexibility of the interpolation scheme. A different proposal was made in \citet{scardapane2016learning}, where the vector $\vect{q} \in \R^T$ of SAF parameters is regularized by penalizing deviations from the values at initialization. Note that it is straightforward to initialize the SAF as any of the known fixed activation functions described before.

%

\subsection{Maxout networks}
\label{sec:maxout_networks}

\begin{figure}
\centering
\includegraphics[width=0.5\columnwidth,keepaspectratio]{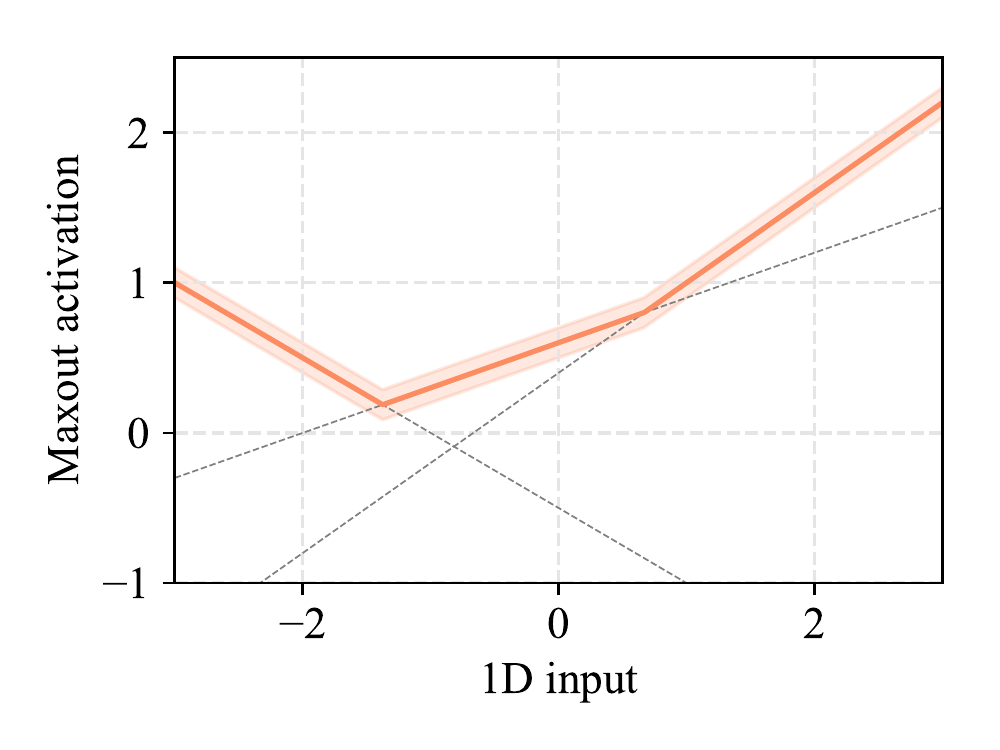} %
\caption{An example of a maxout neuron with a one-dimensional input and $K=3$. The three linear segments are shown with a light gray, while the resulting activation is shown with a shaded red. Note how the maxout can only generate convex shapes by definition. However, plots cannot be made for inputs having more than three dimensions.}
\label{fig:maxout_example}
\end{figure}

Differently from the other functions described up to now, the maxout function introduced in \citet{goodfellow2013maxout} replaces an entire layer in \eqref{eq:nn_layer}. In particular, for each neuron, instead of computing a single dot product $\vect{w}^T\vect{h}$ to obtain the activation (where $\vect{h}$ is the input to the layer), we compute $K$ different products with $K$ separate weight vectors $\vect{w}_1, \ldots, \vect{w}_K$ and biases $b_1, \ldots, b_K$, and take their maximum:
\begin{equation}
g(\vect{h}) = \max_{i=1,\ldots,K} \left\{ \vect{w}_i^T\vect{h} + b_i \right\} \,,
\end{equation}
where the activation function is now a function of a subset of the output of the previous layer. A NN having maxout neurons in all hidden layers is called a maxout network, and remains an universal approximator according to the following theorem.
\begin{theorem}[Theorem 4.3, \citep{goodfellow2013maxout}]
Any continuous function $h(\cdot): \R^{N_0} \rightarrow \R^{N_L}$ can be approximated arbitrarily well on a compact domain by a maxout network with two maxout hidden units, provided $K$ is chosen sufficiently large.
\end{theorem}
The advantage of the maxout function is that it is extremely easy to implement using current linear algebra libraries. However, the resulting functions have several points of non-differentiability, similarly to the APL units. In addition, the number of resulting parameters is generally higher than with alternative formulations. In particular, by increasing $K$ we multiply the original number of parameters by a corresponding factor, while other approaches contribute only linearly to this number. Additionally, we lose the possibility of plotting the resulting activation functions, unless the input to the maxout layer has less than $4$ dimensions. An example with dimension $1$ is shown in Fig. \ref{fig:maxout_example}.

In order to solve the smoothness problem, \citep{zhang2014improving} introduced two smooth versions of the maxout neuron. The first one is the soft-maxout:
\begin{equation}
g(\vect{h}) = \log\left\{ \sum_{i=1}^K \exp\left\{ \vect{w}_i^T\vect{h} + b_i \right\} \right\} \,.
\end{equation}
The second one is the $\ell_p$-maxout, for a user-defined natural number $p$:
\begin{equation}
g(\vect{h}) = \sqrt[p]{ \sum_{i=1}^K \left\lvert \vect{w}_i^T\vect{h} + b_i \right\rvert^p } \,.
\end{equation} 
Closely related to the $\ell_p$-maxout neuron is the $L_p$ unit proposed in \citet{gulcehre2013learned}. Denoting for simplicity $s_i = \vect{w}_i^T\vect{h} + b_i$, the $L_p$ unit is defined as:
\begin{equation}
g(\vect{h}) = \left( \frac{1}{K} \sum_{i=1}^K \lvert s_i - c_i \rvert^{p} \right)^{\frac{1}{p}} \,,
\label{eq:lp_unit}
\end{equation}
where the $K+1$ parameters $\left\{c_1, \ldots, c_K, p\right\}$ are all learned via back-propagation.\footnote{In practice, $p$ is re-parameterized as $1 + \log\left\{1+\exp\left\{p\right\}\right\}$ to guarantee that \eqref{eq:lp_unit} defines a proper norm.} If we fix $c_i=0$, for $p$ going to infinity the $L_p$ unit degenerates to a special case of the maxout neuron:
\begin{equation}
\lim_{p \rightarrow \infty} g(\vect{h}) = \max_{i=1,\ldots,K} \left\{ \lvert s_i \rvert \right\} \,.
\end{equation}

\section{Proposed kernel-based activation functions}
\label{sec:proposed_kernel_afs}

In this section we describe the proposed KAF. Specifically, we model each activation function in terms of a kernel expansion over $D$ terms as:
\begin{equation}
g(s) = \sum_{i=1}^D \alpha_i \kappa\left(s, d_i\right) \,,
\label{eq:proposed_kaf}
\end{equation}
where $\left\{\alpha_i\right\}_{i=1}^D$ are the mixing coefficients, $\left\{d_i\right\}_{i=1}^D$ are the called the dictionary elements, and $\kappa(\cdot, \cdot): \R \times \R \rightarrow \R$ is a 1D kernel function \citep{hofmann2008kernel}. In kernel methods, the dictionary elements are generally selected from the training data. In a stochastic optimization setting, this means that $D$ would grow linearly with the number of training iterations, unless some proper strategy for the selection of the dictionary is implemented \citep{liu2011kernel,van2012kernel}. To simplify our treatment, we consider a simplified case where the dictionary elements are fixed, and we only adapt the mixing coefficients. In particular, we sample $D$ values over the $x$-axis, uniformly around zero, similar to the SAF method, and we leave $D$ as a user-defined hyper-parameter. This has the additional benefit that the resulting model is linear in its adaptable parameters, and can be efficiently implemented for a mini-batch of training data using highly-vectorized linear algebra routines. Note that there is a vast literature on kernel methods with fixed dictionary elements, particularly in the field of Gaussian processes \citep{snelson2006sparse}.

The kernel function $\kappa(\cdot, \cdot)$ needs only respect the positive semi-definiteness property, i.e., for any possible choice of $\left\{\alpha_i\right\}_{i=1}^D$ and $\left\{d_i\right\}_{i=1}^D$ we have that:
\begin{equation}
\sum_{i=1}^D \sum_{j=1}^D \alpha_i \alpha_j \kappa\left(d_i, d_j\right) \ge 0 \,.
\label{eq:psd_kernel}
\end{equation}
For our experiments, we use the 1D Gaussian kernel defined as:
\begin{equation}
\kappa(s, d_i) = \exp\left\{-\gamma\left(s - d_i\right)^2\right\} \,,
\label{eq:gaussian_kernel}
\end{equation}
where $\gamma \in \R$ is called the kernel bandwidth, and its selection is discussed more at length below. Other choices, such as the polynomial kernel with $p \in \mathbb{N}$, are also possible:
\begin{equation}
\kappa(s, d_i) = \left(1 + sd_i\right)^p \,.
\label{eq:polynomial_kernel}
\end{equation}
By the properties of kernel methods, KAFs are equivalent to learning linear functions over a large number of nonlinear transformations of the original activation $s$, without having to explicitly compute such transformations. The Gaussian kernel has an additional benefit: thanks to its definition, the mixing coefficients have only a local effect over the shape of the output function (where the radius depends on $\gamma$, see below), which is advantageous during optimization. In addition, the expression in \eqref{eq:proposed_kaf} with the Gaussian kernel can approximate any continuous function over a subset of the real line \citep{micchelli2006universal}. The expression resembles a one-dimensional radial basis function network, whose universal approximation properties are also well studied \citep{park1991universal}. Below, we go more in depth over some additional considerations for implementing our KAF model. Note that the model has very simple derivatives for back-propagation:
\begin{align}
\frac{\partial g(s)}{\partial \alpha_i} & = \kappa\left(s, d_i\right) \,, \\
\frac{\partial g(s)}{\partial s} & = \sum_{i=1}^{D} \alpha_i \frac{\partial \kappa\left(s, d_i\right)}{\partial s} \,.
\end{align}

\subsection*{On the selection of the kernel bandwidth}

\begin{figure}
\subfloat[$\gamma$ = 2.0]{
\includegraphics[width=0.3\columnwidth,keepaspectratio]{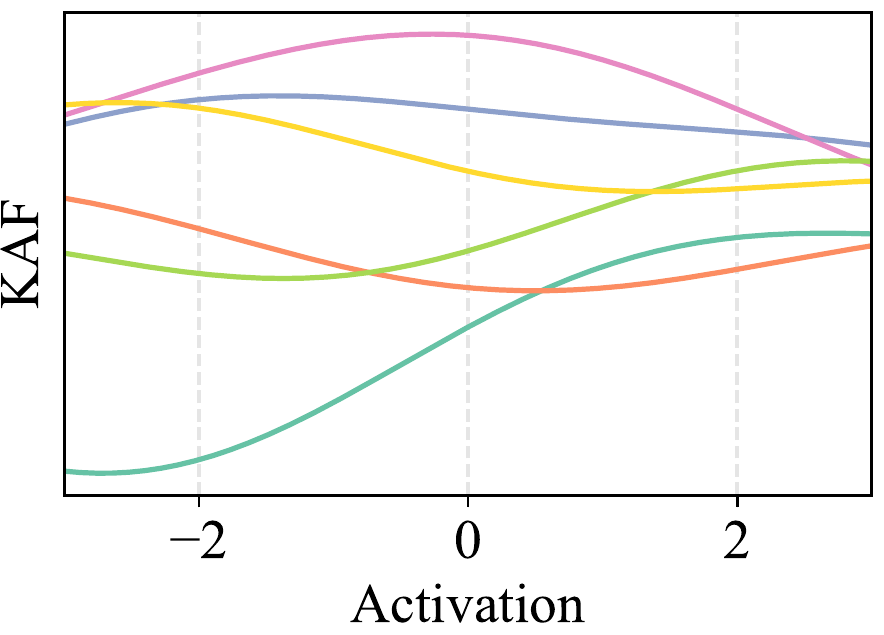}
\label{fig:kaf_example_small_gamma}
} \hfill
\subfloat[$\gamma$ = 0.5]{
\includegraphics[width=0.3\columnwidth,keepaspectratio]{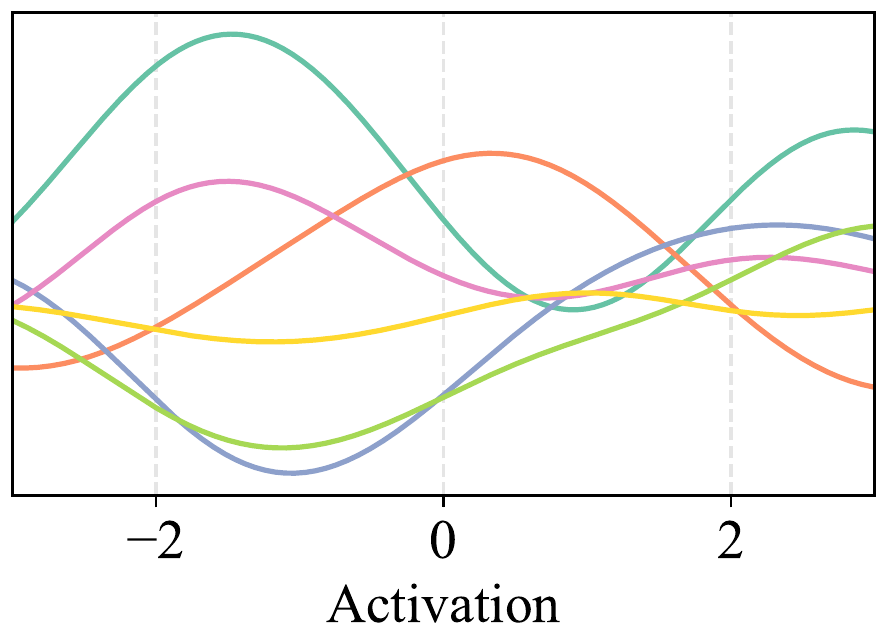}
\label{fig:kaf_example_medium_gamma}
} \hfill
\subfloat[$\gamma$ = 0.1]{
\includegraphics[width=0.3\columnwidth,keepaspectratio]{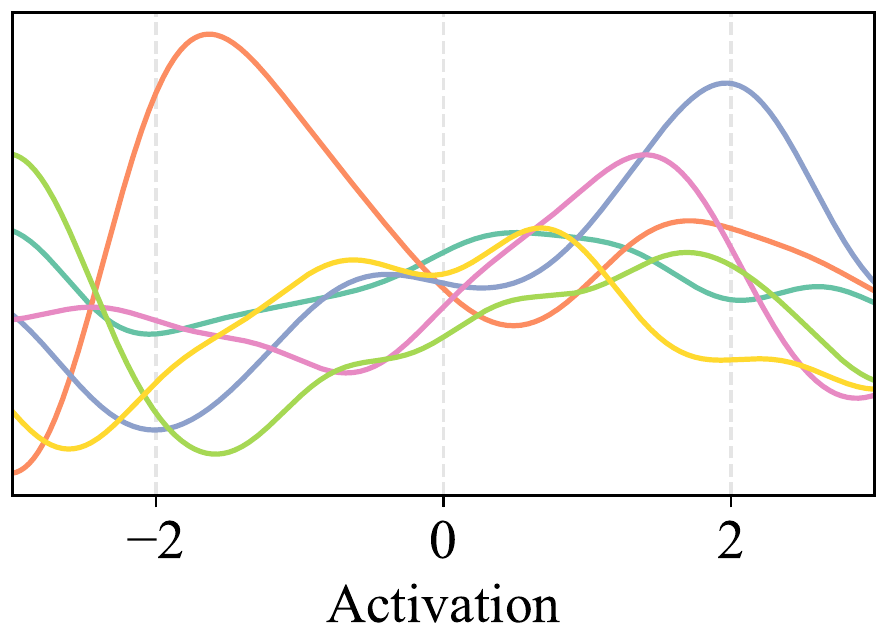}
\label{fig:kaf_example_large_gamma}
} \hfill
\caption{Examples of KAFs. In all cases we sample uniformly $20$ points on the $x$-axis, while the mixing coefficients are sampled from a normal distribution. The three plots show three different choices for $\gamma$.}
\label{fig:kaf_samples}
\end{figure}

Selecting $\gamma$ is crucial for the well-behavedness of the method, by acting indirectly on the effective number of adaptable parameters. In Fig. \ref{fig:kaf_samples} we show some examples of functions obtained by fixing $D=20$, randomly sampling the mixing coefficients, and only varying the kernel bandwidth, showing how $\gamma$ acts on the smoothness of the overall functions.

In the literature, many methods have been proposed to select the bandwidth parameter for performing kernel density estimation \citep{jones1996brief}. These methods include popular rules of thumb such as \citet{scott2015multivariate} or \citet{silverman1986density}.

In the problem of kernel density estimation, the abscissa corresponds to a given dataset with an arbitrary distribution. In the proposed KAF scheme, the abscissa are chosen according to a grid, and as such the optimal bandwidth parameter depends uniquely on the grid resolution. Instead of leaving the bandwidth parameter $\gamma$ as an additional hyper-parameter, we have empirically verified that the following rule of thumb represents a good compromise between smoothness (to allow an accurate approximation of several initialization functions) and flexibility:
\begin{equation}
\gamma = \frac{1}{6\Delta^2} \,,
\label{eq:sigma_rule_of_thumb}
\end{equation}
where $\Delta$ is the distance between the grid points. We also performed some experiments in which $\gamma$ was adapted through back-propagation, though this did not provide any gain in accuracy.


\subsection*{On the initialization of the mixing coefficients}
\label{sec:initialization}

\begin{figure}
\centering
\subfloat[$\tanh$]{
\includegraphics[width=0.45\columnwidth,keepaspectratio]{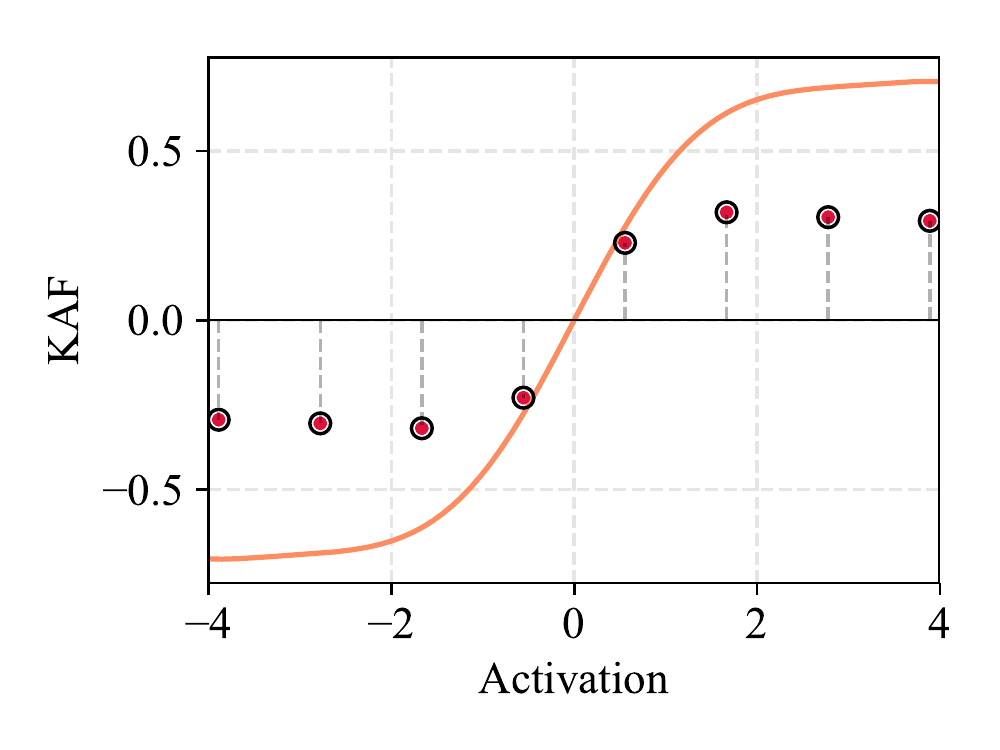}
\label{fig:kaf_initialization_elu}
} \hfill
\subfloat[ELU]{
\includegraphics[width=0.45\columnwidth,keepaspectratio]{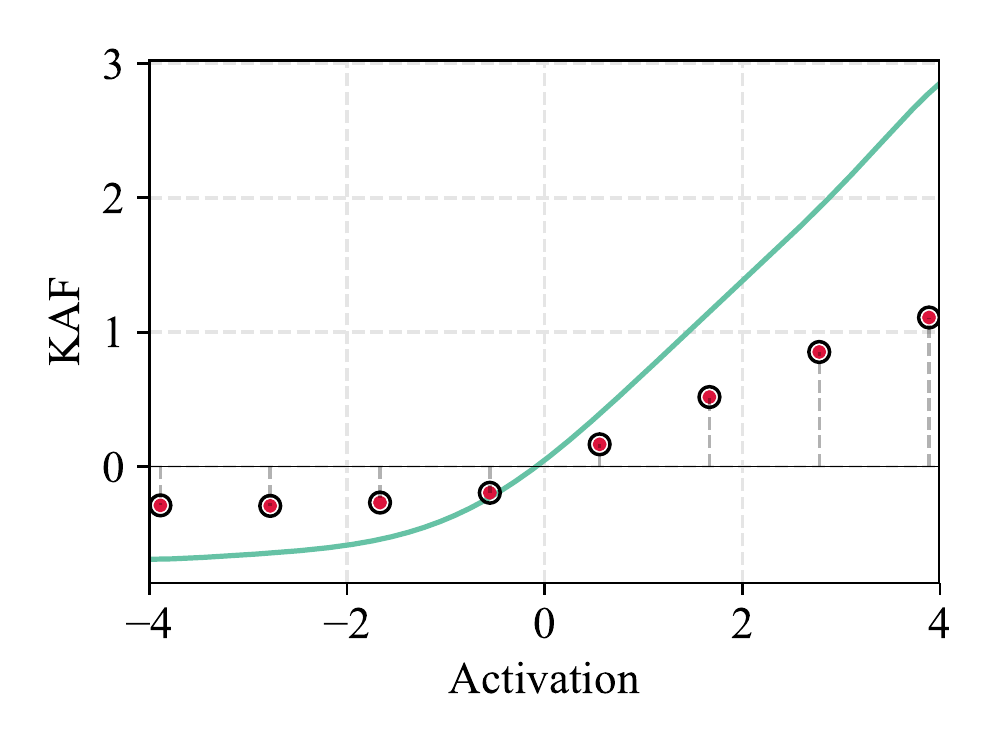}
\label{fig:kaf_initialization_tanh}
} \hfill
\caption{Two examples of initializing a KAF using \eqref{eq:kaf_initialization_krr}, with $\varepsilon=10^{-6}$. (a) A hyperbolic tangent. (b) The ELU in \eqref{eq:elu}. The red dots indicate the corresponding initialized values for the mixing coefficients.}
\label{fig:kaf_initialization_examples}
\end{figure}

A random initialization of the mixing coefficients from a normal distribution, as in Fig. \ref{fig:kaf_samples}, provides good diversity for the optimization process. Nonetheless, a further advantage of our scheme is that we can initialize some (or all) of the KAFs to follow any know activation function, so as to guarantee a certain desired behavior. Specifically, denote by $\vect{t} = \left[t_1, \ldots, t_D\right]^T$ the vector of desired initial KAF values corresponding to the dictionary elements $\vect{d} = \left[d_1, \ldots, d_D\right]^T$. We can initialize the mixing coefficients $\boldsymbol{\alpha} = \left[\alpha_1, \ldots, \alpha_D\right]^T$ using kernel ridge regression:
\begin{equation}
\boldsymbol{\alpha} = \left(\vect{K} + \varepsilon\vect{I}\right)^{-1}\vect{t} \,,
\label{eq:kaf_initialization_krr}
\end{equation}
where $\vect{K} \in \R^{D \times D}$ is the kernel matrix computed between $\vect{t}$ and $\vect{d}$, and we add a diagonal term with $\varepsilon > 0$ to avoid degenerate solutions with very large mixing coefficients. Two examples are shown in Fig. \ref{fig:kaf_initialization_examples}.

\subsection*{Multi-dimensional kernel activation functions}

In our experiments, we also consider a two-dimensional variant of the proposed KAF, that we denote as 2D-KAF. Roughly speaking, the 2D-KAF acts on a pair of activation values, instead of a single one, and learns a two-dimensional function to combine them. It can be seen as a generalization of a two-dimensional maxout neuron, which is instead constrained to output the maximum value among the two inputs.

Similarly to before, we construct a dictionary $\vect{d} \in \R^{D^2 \times 2}$ by sampling a uniform grid over the 2D plane, by considering $D$ positions uniformly spaced around $0$ in both dimensions. We group the incoming activation values in pairs (assuming that the layer has even size), and for each possible pair of activations $\vect{s} = \left[s_{k}, s_{k+1}\right]^T$ we output:
\begin{equation}
g\left(\vect{s}\right) = \sum_{i=1}^{D^2} \alpha_i \kappa\left(\vect{s}, \vect{d}_i\right) \,,
\label{eq:2d_kaf}
\end{equation}
where $\vect{d}_i$ is the $i$-th element of the dictionary, and we now have $D^2$ adaptable coefficients $\left\{\alpha_i\right\}_{i=1}^{D^2}$. In this case, we consider the 2D Gaussian kernel:
\begin{equation}
\kappa\left(\vect{s}, \vect{d}_i\right) = \exp\left\{ -\gamma\norm{\vect{s} - \vect{d}_i}^2 \right\} \,,
\label{eq:2d_gaussian_kernel}
\end{equation}
where we use the same rule of thumb in \eqref{eq:sigma_rule_of_thumb}, multiplied by $\sqrt{2}$, to select $\gamma$. The increase in parameters is counter-balanced by two factors. Firstly, by grouping the activations we halve the size of the linear matrix in the subsequent layer. Secondly, we generally choose a smaller $D$ with respect to the $1D$ case, i.e., we have found that values in $\left[5, 10\right]$ are enough to provide a good degree of flexibility. Table \ref{tab:non_parametric_afs_comparison} provides a comparison of the two proposed KAF models to the three alternative non-parametric activation functions described before. We briefly mention here that a multidimensional variant of the SAF was explored in \citet{solazzi2000artificial}.

\begin{sidewaystable}
\begin{threeparttable}
{\centering\hfill{}
\setlength{\tabcolsep}{4pt}
\renewcommand{\arraystretch}{1.4}
\begin{footnotesize}
\begin{tabular}{lcccccc}   
\toprule
\textbf{Name} & \textbf{Smooth} & \textbf{Locality} & \textbf{Can use regularization} & \textbf{Plottable} & \textbf{Hyper-parameter} & \textbf{Trainable weights} \\
\midrule
APL & No & Partially & Only $\ell_2$ regularization & Yes & Number of segments $S$ & $N_{i-1}N_{i} + N_{i} + 2SN_{i}$ \\
SAF & Yes & Yes & No & Yes & Number of control points $Q$ & $N_{i-1}N_{i} + N_{i} + QN_{i}$ \\
Maxout & No & No & Yes & No\tnote{*} & Number of affine maps $K$ & $KN_{i-1}N_{i} + KN_{i}$ \\
\midrule
\textbf{Proposed KAF} & Yes & Yes\tnote{**} & Yes & Yes & Size of the dictionary $D$ & $N_{i-1}N_{i} + N_{i} + DN_{i}$ \\
\textbf{Proposed 2D-KAF} & Yes & Yes\tnote{**} & Yes & Yes & Size of the dictionary $D$ & $N_{i-1}N_{i} + \frac{\left(N_{i} + D^2N_{i}\right)}{2}$ \\
\bottomrule
\end{tabular}
\end{footnotesize}
}
\hfill{}
\begin{tablenotes}
\item[*] Maxout functions can only be plotted whenever $N_{i-1} \le 3$, which is almost never the case.
\item[**] Only when using the Gaussian (or similar) kernel function.
\end{tablenotes}
\end{threeparttable}
\caption{A comparison of the existing non-parametric activation functions and the proposed KAF and 2D-KAF. In our definition, an activation function is local if each adaptable weight only affects a small portion of the output values.}
\label{tab:non_parametric_afs_comparison}
\end{sidewaystable}

\section{Related work}
\label{sec:related_work}

Many authors have considered ways of improving the performance of the classical activation functions, which do not necessarily require to adapt their shape via numerical optimization, or that require special care when implemented. For completeness, we briefly review them here before moving to the experimental section.

As stated before, the problem of ReLU is that its gradient is zero outside the `active' regime where $s \ge 0$. To solve this, the randomized leaky ReLU \citep{xu2015empirical} considers a leaky ReLU like in \eqref{eq:leaky_relu}, in which during training the parameter $\alpha$ is randomly sampled at every step in the uniform distribution $\mathcal{U}(l,u)$, and the lower/upper bounds $\left\{l, u \right\}$ are selected beforehand. To compensate with the stochasticity in training, in the test phase $\alpha$ is set equal to:
\begin{equation}
\alpha = \frac{l + u}{2} \,,
\end{equation}
which is equivalent to taking the average of all possible values seen during training. This is similar to the dropout technique \citep{srivastava2014dropout}, which randomly deactivates some neurons during each step of training, and later rescales the weights during the test phase.

\begin{figure}
\subfloat[Original function]{
\includegraphics[width=0.3\columnwidth,keepaspectratio]{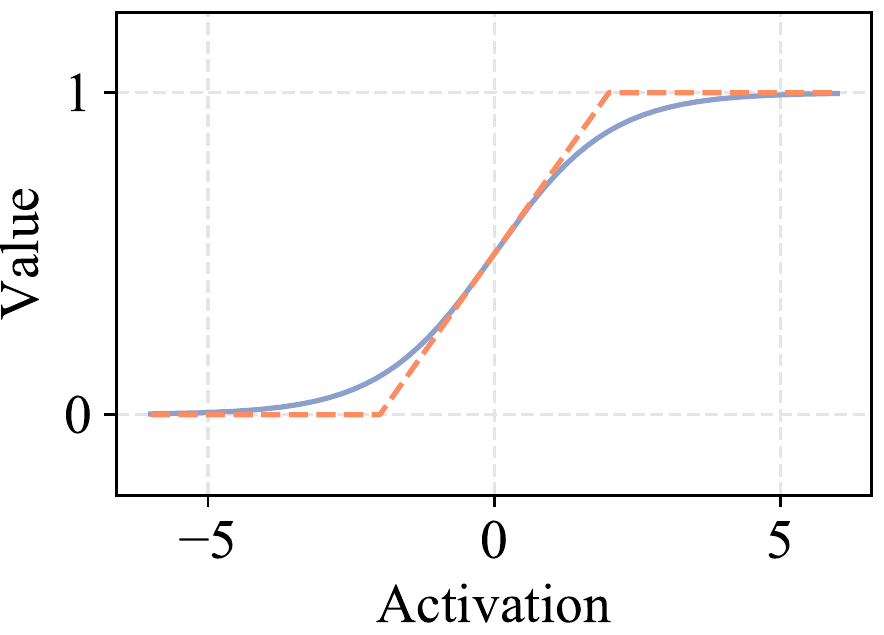}
\label{fig:hard_thresholded_sigmoid}
} \hfill
\subfloat[Noise]{
\includegraphics[width=0.3\columnwidth,keepaspectratio]{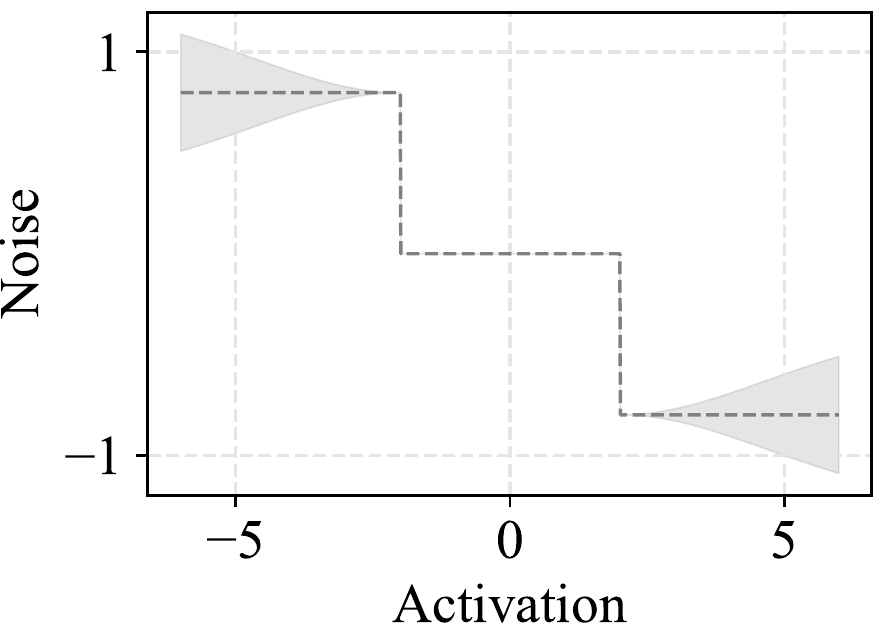}
\label{fig:noise_variance}
} \hfill
\subfloat[Noisy activation function]{
\includegraphics[width=0.3\columnwidth,keepaspectratio]{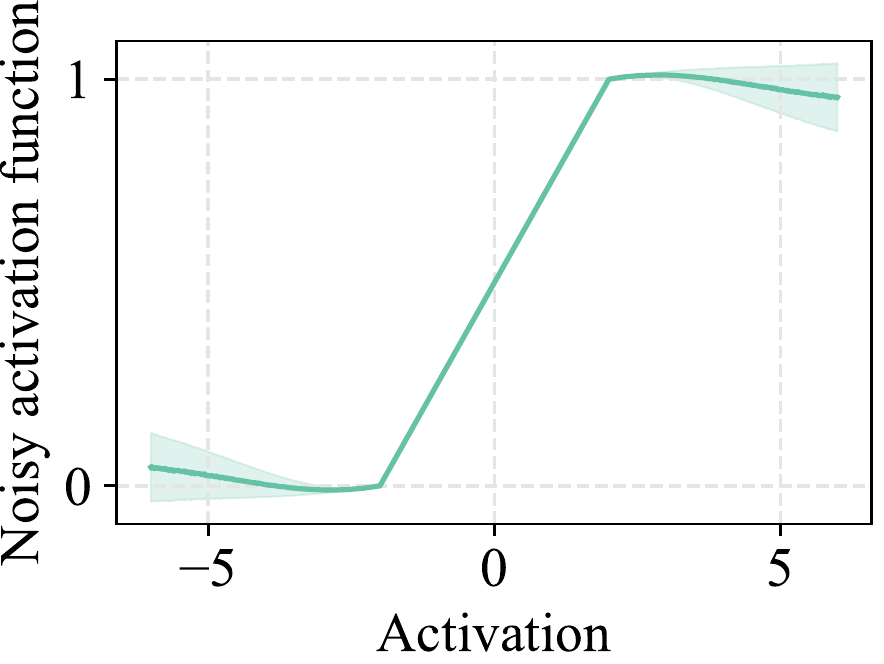}
\label{fig:noisy_sigmoid_af}
} \hfill
\caption{An example of noisy activation function. (a) Original sigmoid function (blue), together with its hard-thresholded version (in red). (b) For any possible activation value outside the saturated regime, we add random half-normal noise with increasing variance and matching sign according to the algorithm in \citet{gulcehre2016noisy} (the shaded areas correspond to one standard deviation). (c) Final noisy activation function computed as in \citet{gulcehre2016noisy}.  At test time, only the expected values (represented with solid green line) are returned.}
\label{fig:noisy_af}
\end{figure}

More in general, several papers have developed stochastic versions of the classical artificial neurons, whose output depend on one or more random variables sampled during their execution \citep{bengio2013estimating}, under the idea that the resulting noise can help guide the optimization process towards better minima. Notably, this provides a link between classical NNs and other probabilistic methods, such as generative networks and networks trained using variational inference \citep{bengio2014deep,schulman2015gradient}. The main challenge is to design stochastic neurons that provide a simple mechanism for back-propagating the error through the random variables, without requiring expensive sampling procedures, and with a minimal amount of interference over the network. As a representative example, the noisy activation functions proposed in \citet{gulcehre2016noisy} achieve this by combining activation functions with `hard saturating' regimes (i.e., their value is exactly zero outside a limited range) with random noise over the outputs, whose variance increases in the regime where the function saturates to avoid problems due to the sparse gradient terms.  An example is given in Fig. \ref{fig:noisy_af}.

%
%
%
%

Another approach is to design \textit{vector-valued} activation functions to maximize parameter sharing. In the simplest case, the concatenated ReLU \citep{shang2016understanding} returns two output values by applying a ReLU function both on $s$ and on $-s$. Similarly, the order statistic network \citep{rennie2014deep} modifies a maxout neuron by returning the input activations in sorted order, instead of picking the highest value only. Multi-bias activation functions \citep{li2016multi} compute several activations values by using different bias terms, and then apply the same activation function independently over each of the resulting values. The network-in-network \citep{lin2013network} model is a non-parametric approach specific to convolutional neural networks, wherein the nonlinear units are replaced with a fully connected NN.

For specific tasks of audio modeling, some authors have proposed the use of Hermite polynomials for adapting the activation functions \citep{siniscalchi2013hermitian,siniscalchi2017adaptation}. Similarly to our proposed KAF, the functions are expressed as a weighted sum of several fixed nonlinear transformations of the activation values, i.e., the Hermite polynomials. However, the nonlinear transformations are computed through the use of a recurrence formula, thus highly increasing the computational load.

\section{Experimental results}
\label{sec:experiments}

In this section we provide a comprehensive evaluation of the proposed KAFs and 2D-KAFs when applied to several use cases. As a preliminary experiment, we begin by comparing multiple activation functions on a relatively small classification dataset (Sensorless) in Section \ref{sec:visualizing_functions}, where we discuss several examples of the shapes that are generally obtained by the networks and initialization strategies. We then consider a large-scale dataset taken from \citet{baldi2014searching} in Section \ref{sec:results_susy}, where we show that two layers of KAFs are able to significantly outperform a feedforward network with five hidden layers, even when considering parametric activation functions and state-of-the-art regularization techniques. In Section \ref{sec:results_cifar10} we show that KAFs and 2D-KAF provide an increase in performance also when applied to convolutional layers on the CIFAR-10 dataset. Finally, we show in Section \ref{sec:results_rl} that they have significantly faster training and higher cumulative reward for a set of reinforcement learning scenario using MuJoCo environments from the OpenAI Gym\footnote{\url{https://gym.openai.com/}}. We provide an open-source library to replicate the experiments, with the implementation of KAFs and 2D-KAFs in three separate frameworks, i.e., AutoGrad\footnote{\url{https://github.com/HIPS/autograd}}, TensorFlow\footnote{\url{https://www.tensorflow.org/}}, and PyTorch\footnote{\url{http://pytorch.org/}}, which is publicly accessible on the web\footnote{\url{https://github.com/ispamm/kernel-activation-functions/}}.

\subsection{Experimental setup}
Unless noted otherwise, in all experiments we linearly preprocess the input features between -1 and +1, and we substitute eventual missing values with the median values computed from the corresponding feature columns. From the full dataset, we randomly keep a portion of the dataset for validation and another portion for test. All neural networks use a softmax activation function in their output layer, and they are trained by minimizing the average cross-entropy on the training dataset, to which we add a small $\ell_2$-regularization term whose weight is selected in accordance to the literature. For optimization, we use the Adam algorithm \citep{kingma2014adam} with mini-batches of 100 elements and default hyper-parameters. For each epoch we compute the accuracy over the validation set, and we stop training whenever the validation accuracy is not improving for 15 consecutive epochs. Experiments are performed using the PyTorch implementation on a machine with an Intel Xeon E5-2620 CPU, with 16 GB of RAM and a CUDA back-end employing an Nvidia Tesla K20c. All accuracy measures over the test set are computed by repeating the experiments for 5 different splits of the dataset (unless the splits are provided by the dataset itself) and initializations of the networks. Weights of the linear layers are always initialized using the so-called `Uniform He' strategy, while additional parameters introduced by parametric and non-parametric activation functions are initialized by following the guidelines of the original papers.

\subsection{Visualizing the activation functions}
\label{sec:visualizing_functions}

We begin with an experiment on the `Sensorless' dataset to investigate whether KAFs and 2D-KAFs can indeed provide improvements in accuracy with respect to other baselines, and for visualizing some of the common shapes that are obtained after training. The Sensorless dataset is a standard benchmark for supervised techniques, composed of 58509 examples with 49 input features representing electric signals, that are used to predict one among 11 different classes representing operating conditions. We partition it using a random 15\% for validation and another 15\% for testing, and we use a small regularization factor of $10^{-4}$.

In this dataset, we found that the best performing fixed activation function is a simple hyperbolic tangent. In particular, a network with one hidden layers of 100 neurons achieves a test accuracy of $97.75 \%$, while the best result is obtained with a network of three hidden layers (each composed of 100 neurons), which achieves a test accuracy of $99.18\%$. Due to the simplicity of the dataset, we have not found improvements here by adding more layers or including dropout during training as in the following sections. The best performing parametric activation function is instead the PReLU, that improves the results by obtaining a $98.48\%$ accuracy with a single hidden layer, and $99.30\%$ with three hidden layers.

\begin{figure}
\subfloat[]{
\includegraphics[width=0.28\columnwidth,keepaspectratio]{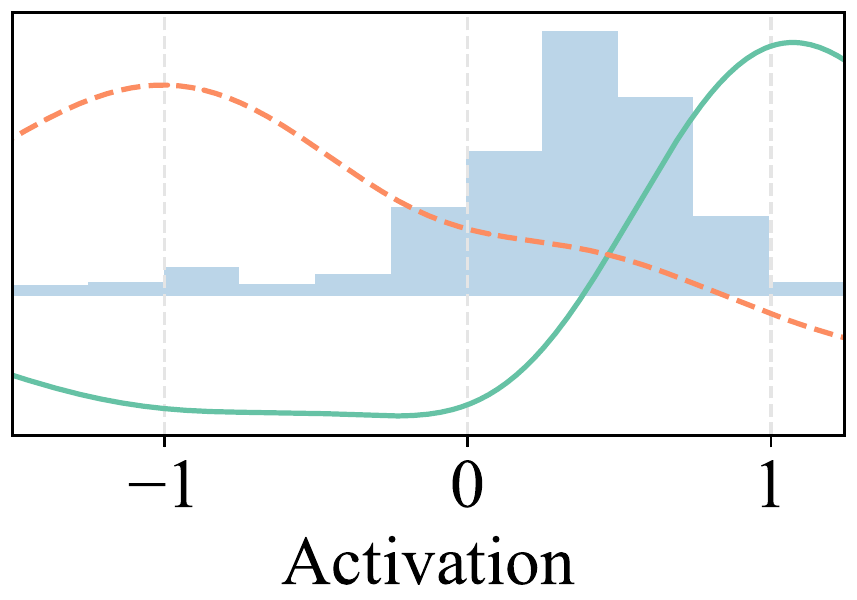}
\label{fig:kaf_examples_0}
} \hfill
\subfloat[]{
\includegraphics[width=0.28\columnwidth,keepaspectratio]{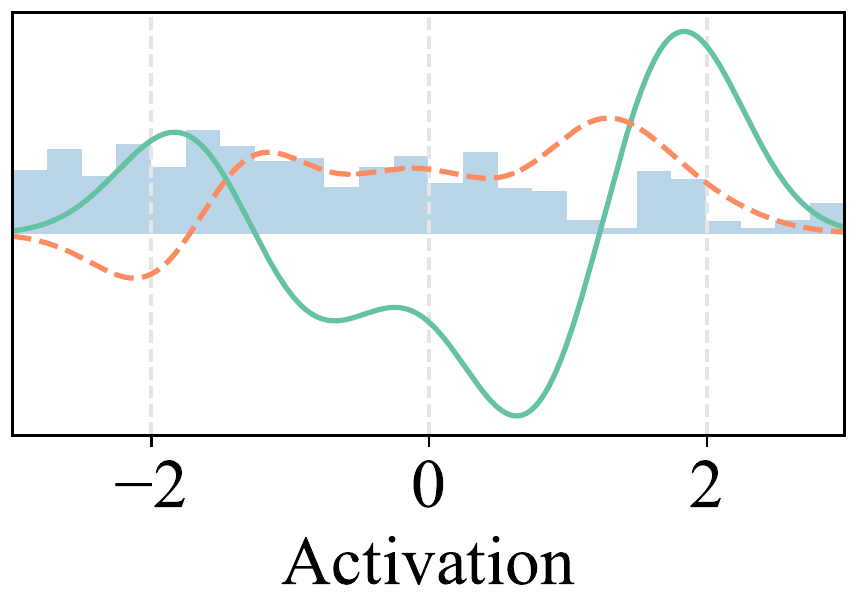}
\label{fig:kaf_examples_1}
} \hfill
\subfloat[]{
\includegraphics[width=0.28\columnwidth,keepaspectratio]{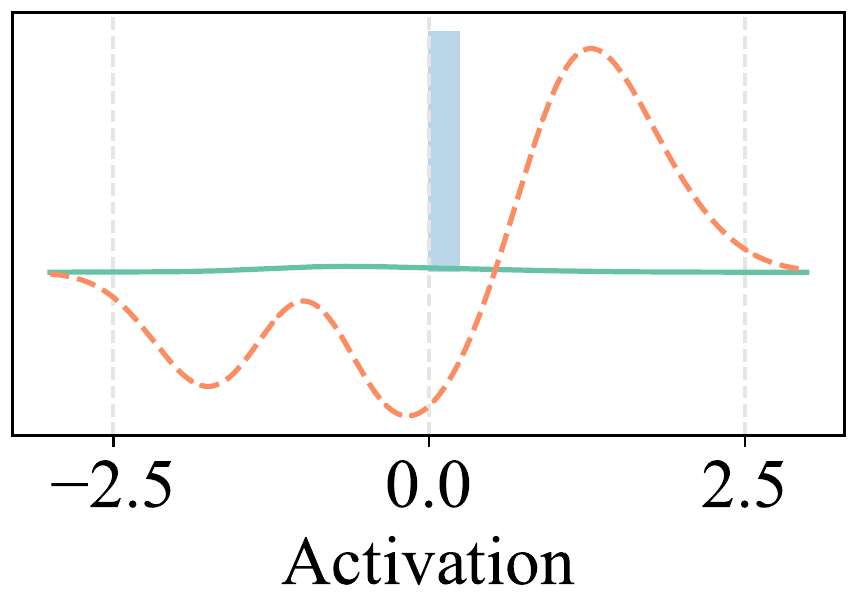}
\label{fig:kaf_examples_5}
} \vfill
\subfloat[]{
\includegraphics[width=0.28\columnwidth,keepaspectratio]{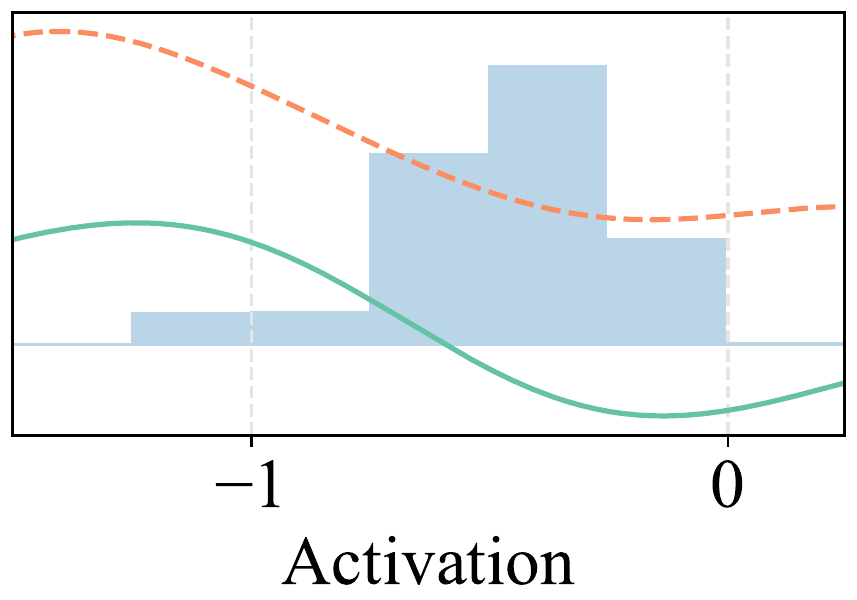}
\label{fig:kaf_examples_6}
} \hfill
\subfloat[]{
\includegraphics[width=0.28\columnwidth,keepaspectratio]{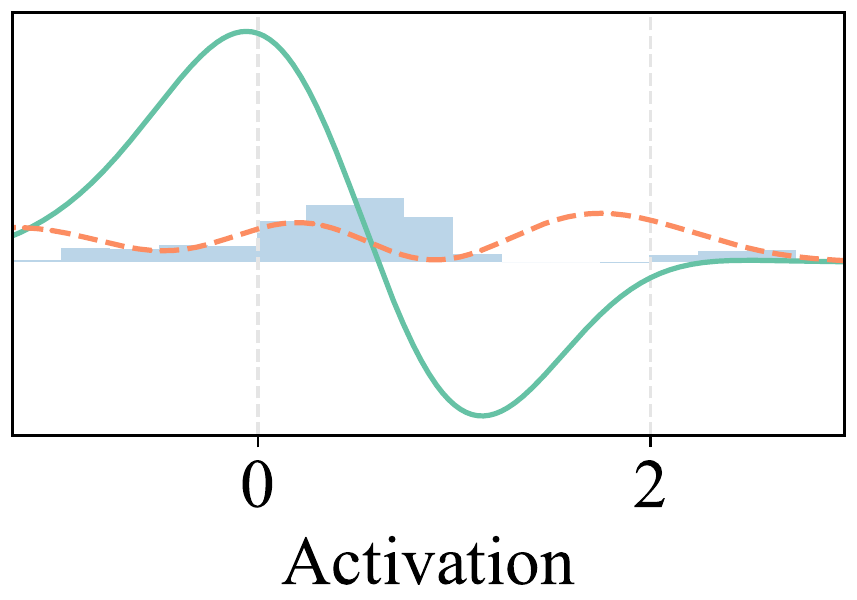}
\label{fig:kaf_examples_18}
} \hfill
\subfloat[]{
\includegraphics[width=0.28\columnwidth,keepaspectratio]{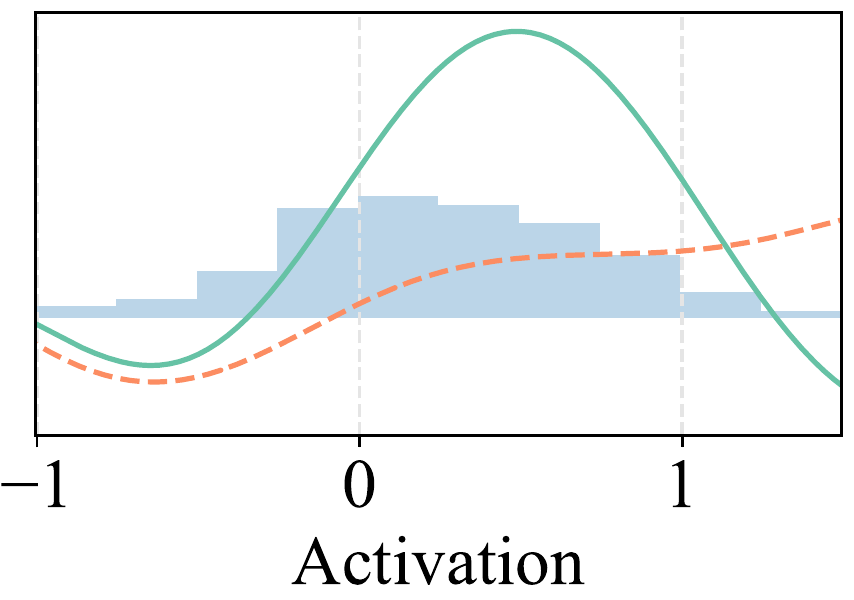}
\label{fig:kaf_examples_46}
} \vfill
\caption{Examples of $6$ trained KAFs (with random initialization) on the Sensorless dataset. On the $y$-axis we plot the output value of the KAF. The KAF after initialization is shown with a dashed red, while the final KAF is shown with a solid green. The distribution of activation values after training is shown as a reference with a light blue.}
\label{fig:kaf_examples_random_initialization}
\end{figure}

For comparison, we train several feedforward networks with KAFs in the hidden layers, having a dictionary of $D=20$ elements equispaced between $-3.0$ and $3.0$, and initializing the linear coefficients from a normal distribution with mean $0$ and variance $0.3$. Using this setup, we already outperform all other baselines obtaining an accuracy of $99.04\%$ with a single hidden layer, which improves to $99.80\%$ when considering two hidden layers of KAFs.

Although the dataset is relatively simple, the shapes we obtain are representative of all the experiments we performed, and we provide a selection in Fig. \ref{fig:kaf_examples_random_initialization}. Specifically, the initialization of the KAF is shown with a dashed red line, while the final KAF is shown with a solid green line. For understanding the behavior of the functions, we also plot the empirical distribution of the activations on the test set using a light blue in the background. Some shapes are similar to common activation functions discussed in Section \ref{sec:fixed_af}, although they are shifted on the $x$-axis to correspond to the distribution of activation values. For example Fig. \ref{fig:kaf_examples_0} is similar to an ELU, while Fig. \ref{fig:kaf_examples_6} is similar to a standard saturating function. Also, while in the latter case the final shape is somewhat determined by the initialization, the final shapes in general tend to be independent from initialization, as in the case of Fig. \ref{fig:kaf_examples_0}. Another common shape is that of a radial-basis function, as in Fig. \ref{fig:kaf_examples_46}, which is similar to a Gaussian function centered on the mean of the empirical distribution. Shapes, however, can be vastly more complex than these. For example, in Fig. \ref{fig:kaf_examples_18} we show a function which acts as a standard saturating function on the main part of the activations' distribution, while its right-tail tends to remove values larger than a given threshold, effectively acting as a sort of implicit regularizer. In Fig. \ref{fig:kaf_examples_1} we show a KAF without an intuitive shape, that selectively amplify (either positively or negatively) multiple regions of its activation space. Finally, in Fig. \ref{fig:kaf_examples_5} we show an interesting pruning effect, where useless neurons correspond to activation functions that are practically zero everywhere. This, combined with the possibility of applying $\ell_1$-regularization \citep{scardapane2017group} allows to obtain networks with a significant smaller number of effective parameters. 

Interestingly, the shapes obtained in Fig. \ref{fig:kaf_examples_random_initialization} seem to be necessary for the high performance of the networks, and they are not an artifact of initialization. Specifically, we obtain similar accuracies (and similar output functions) even when initializing all KAFs as close as possible to hyperbolic tangents, following the method described in Section \ref{sec:initialization}, while we obtain a vastly inferior performance (in some cases even worse than the baseline), if we initialize the KAFs randomly and we prevent their adaptation. This (informally) points to the fact that their flexibility and adaptability seems to be an intrinsic component of their good performance in this experiment and in the following sections, an aspect that we will return to in our conclusive section.

\begin{figure}
\begin{minipage}{0.6cm}
\vspace{-1.5\columnwidth}\rotatebox{90}{{\footnotesize Activation 2}}
\end{minipage}%
\begin{minipage}{\dimexpr\linewidth-2.50cm\relax}%
  \raisebox{\dimexpr-.5\height-1em}{\includegraphics[scale=0.36]{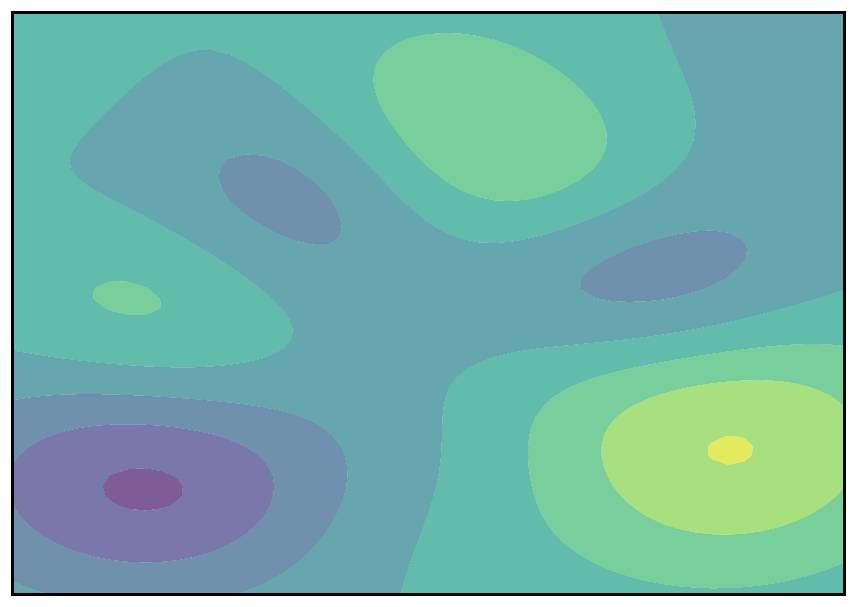}\hspace{0.2em}\includegraphics[scale=0.36]{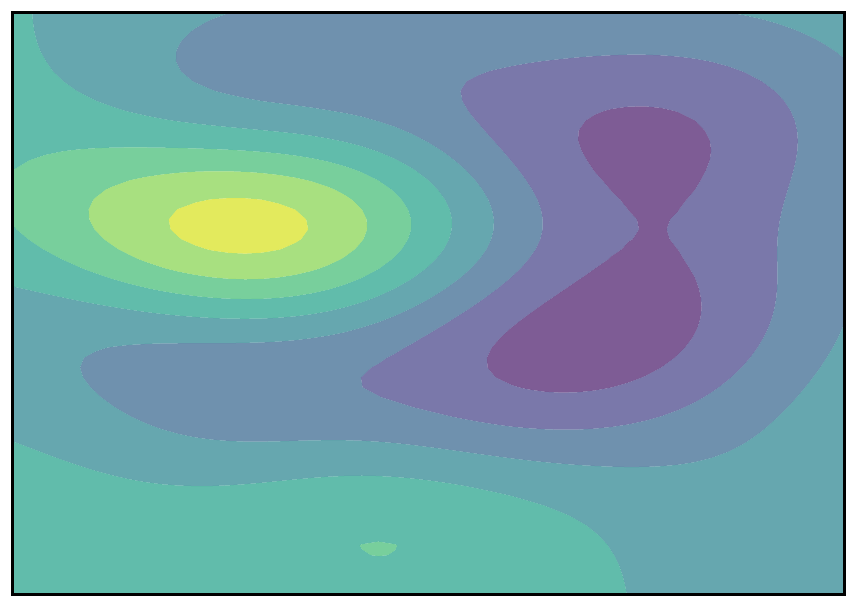}\hspace{0.2em}\includegraphics[scale=0.36]{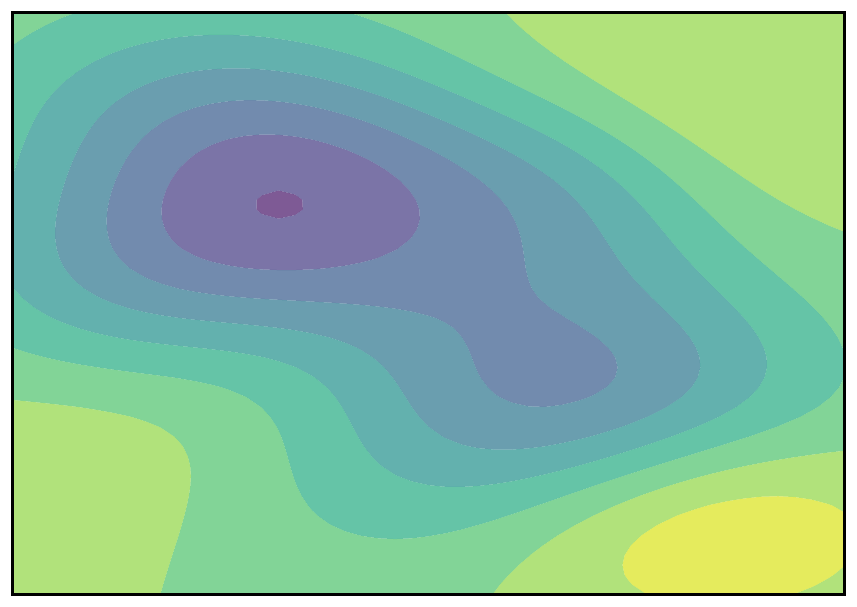}\hspace{0.2em}\includegraphics[scale=0.36]{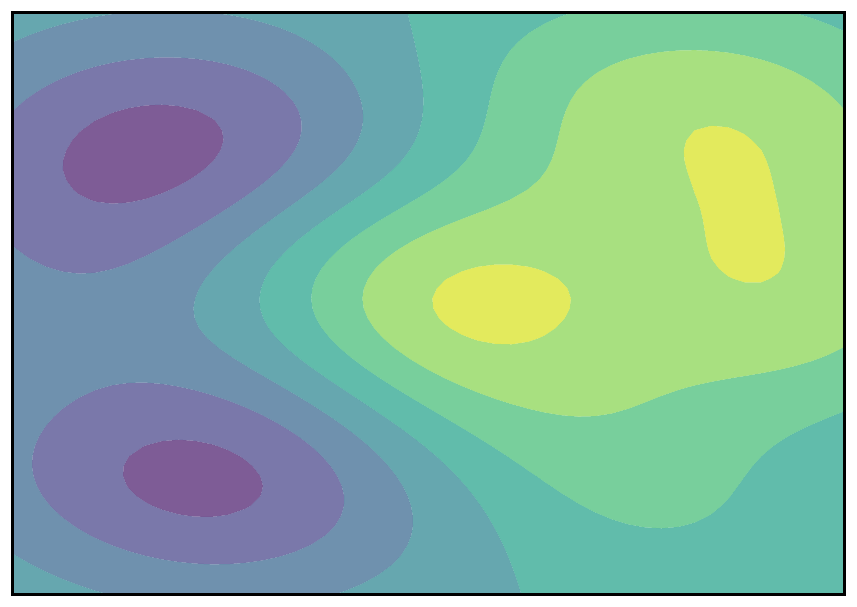}}\ {} \\ 
  \raisebox{\dimexpr-.5\height-1em}{\includegraphics[scale=0.36]{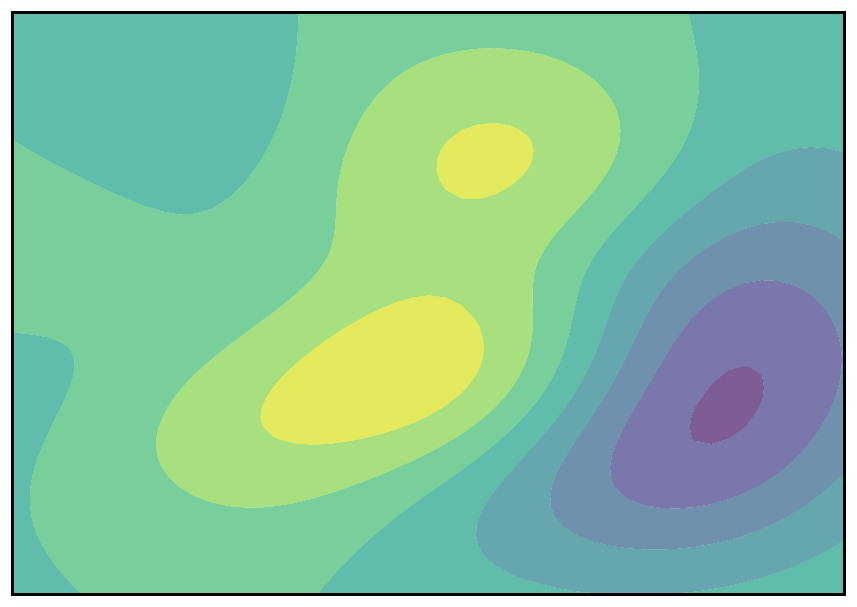}\hspace{0.2em}\includegraphics[scale=0.36]{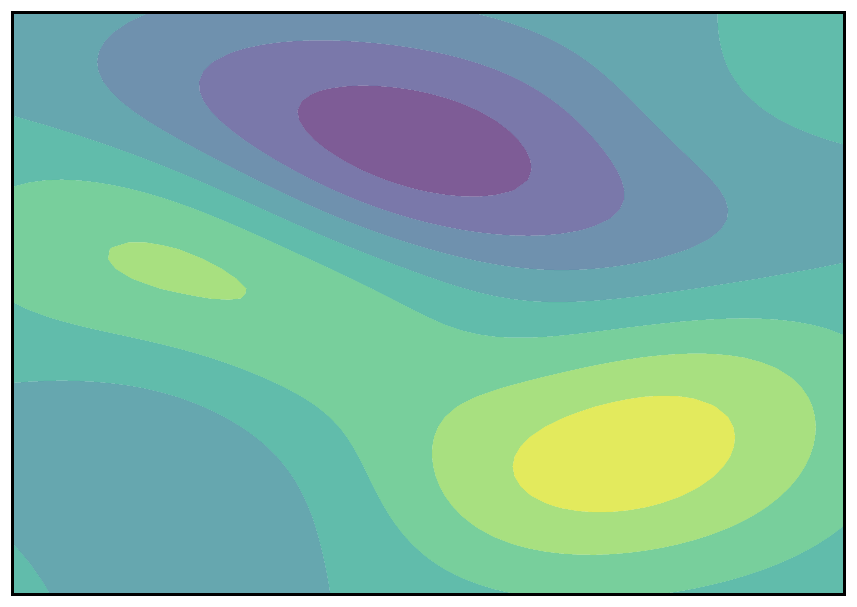}\hspace{0.2em}\includegraphics[scale=0.36]{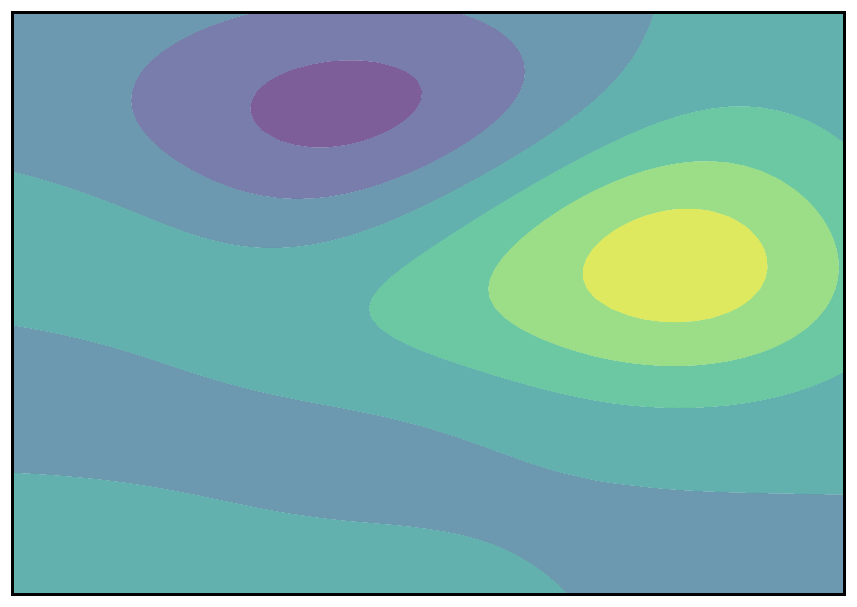}\hspace{0.2em}\includegraphics[scale=0.36]{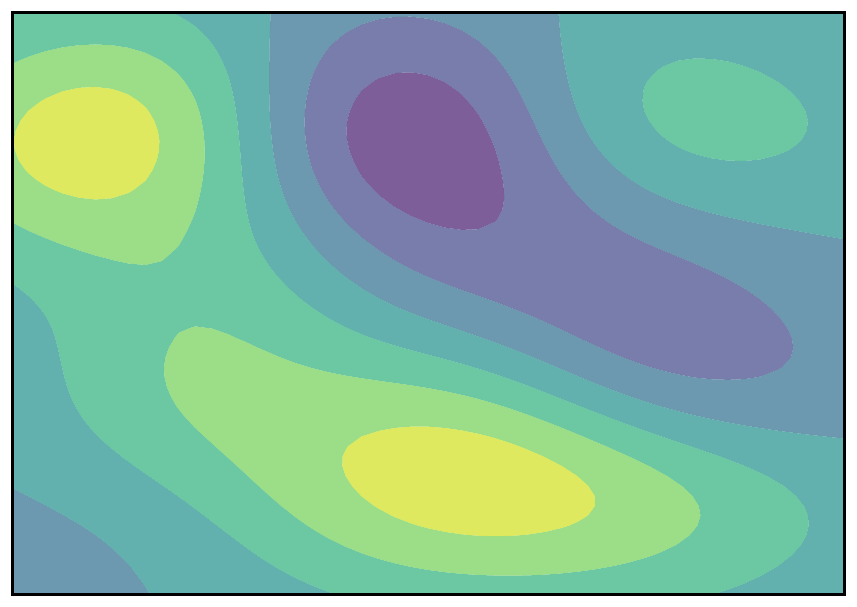}}\ {}

  \vspace*{0.1cm}\hspace*{0.485\columnwidth}{\footnotesize Activation 1}
 \end{minipage}%
  \caption{Examples of $8$ trained 2D-KAFs on the Sensorless dataset. On the $x$- and $y$-axis we plot the two activation values (ranging from $-3$ to $3$), while we show the output of the KAF using a heat map, where darker colors represent larger output values and lighter colors represent lower activation values, respectively.}
  \label{fig:2dkaf_examples}
\end{figure}

Results are also similar when using 2D-KAFs, that we initialize with $D=10$ elements on each axis using the same strategy as for KAFs. In this scenario, they obtain a test accuracy of $98.90\%$ with a single hidden layer, and an accuracy of $99.84\%$ (thus improving over the KAF) for two hidden layers. Some examples of obtained shapes are provided in Fig. \ref{fig:2dkaf_examples}.

\subsection{Comparisons on the SUSY benchmark}
\label{sec:results_susy}

In this section, we evaluate the algorithms on a realistic large-scale use case, the SUSY benchmark introduced in \citet{baldi2014searching}. The task is to predict the presence of super symmetric particles on a simulation of a collision experiment, starting from 18 features (both low-level and high-level) describing the simulation itself. The overall dataset is composed of five million examples, of which the last 500000 are used for test, and another 500000 for validation. The task is interesting for several reasons. Due to the nature of the data, even a tiny change in accuracy (measured in terms of area under the curve, AUC) is generally statistically significant. In the original paper, \citet{baldi2014searching} showed the best AUC was obtained by a deep feedforward network having five hidden layers, with significantly better results when compared to a shallow network. Surprisingly, \citet{agostinelli2014learning} later showed that a shallow network is in fact sufficient, so long as it uses non-parametric activation functions (in that case, APL units).

In order to replicate these results with our proposed methods, we consider a baseline network inspired to \citet{baldi2014searching}, having five hidden layers with 300 neurons each and ReLU activation functions, with dropout applied to the last two hidden layer with probability 0.5. For comparison, we also consider the same architecture, but we substitute ReLUs with ELU, SELU, and PReLU functions. For SELU, we also substitute the standard dropout with the customized version proposed in \citet{klambauer2017self}. We compare with simpler networks composed of one or two hidden layers of 300 neurons, employing Maxout neurons (with $K=5$), APL units (with $S=3$ as proposed in \citet{agostinelli2014learning}), and the proposed KAFs and 2D-KAFs, following the same random initializations as the previous section. Results, in terms of AUC and amount of trainable parameters, are given in Table \ref{tab:results_susy}.

\begin{table}
{\centering\hfill{}
	\setlength{\tabcolsep}{4pt}
	\renewcommand{\arraystretch}{1.5}
	\begin{footnotesize}
	\begin{tabular}{lcc}   
	\toprule
	\textbf{Activation function} & \textbf{Testing AUC} & \textbf{Trainable parameters}\\ 
	\midrule
	ReLU & $0.8739 (0.001)$ & \multirow{3}{*}{$367201$} \\
	ELU & $0.8739 (0.001)$ & \\
	SELU & $0.8745 (0.002)$ & \\
	\midrule
	PReLU & $0.8748 (0.001)$ & $368701$ \\
	\midrule
	Maxout (one layer) & $0.8744 (0.001)$ & $17401$ \\
	Maxout (two layers) & $0.8744 (0.002)$ & $288301$ \\
	\midrule
	APL (one layer) & $0.8744 (0.002)$ & $7801$ \\
	APL (two layers) & $0.8757 (0.002)$ & $99901$ \\
	\midrule
	KAF (one layer) & $0.8756 (0.001)$ & $12001$ \\
	KAF (two layers) & $\underline{0.8758 (0.001)}$ & $108301$ \\
	\midrule
	2D-KAF (one layer) & $0.8750 (0.001)$ & $20851$ \\
	2D-KAF (two layers) & $\vect{0.8764 (0.002)}$ & $81151$ \\
	\bottomrule
	\end{tabular}
	\end{footnotesize}
}
\hfill{}
\caption{Results of different activation functions on the SUSY benchmark. The last four rows are the proposed KAF and 2D-KAF. Standard deviation for the AUC is given between brackets, the best result is shown in bold, and the second best result is underlined. All networks with fixed or parametric activation functions have five hidden layers. See the text for a full description of the architectures.}
\label{tab:results_susy}
\end{table}

There are several clear results that emerge from the analysis of Table \ref{tab:results_susy}. First of all, the use of sophisticated activation functions (such as the SELU), or of parametric functions (such as the PReLU) can improve performance, in some cases even significantly. However, these improvements still require several layers of depth, while they both fail to provide accurate results when experimenting with shallow networks. On the other hand, all non-parametric functions are able to achieve similar (or even superior) results, while only requiring one or two hidden layers of neurons. Among them, APL and Maxout achieve a similar AUC with one layer, but only APL is able to benefit from the addition of a second layer. Both KAF and 2D-KAF are able to significantly outperform all the competitors, and the overall best result is obtained by a 2D-KAF network with two hidden layers. This is obtained with a significant reduction in the number of trainable parameters, as also described more in depth in the following section.

\subsection{Experiments with convolutive layers on CIFAR-10}
\label{sec:results_cifar10}

Although our focus has been on feedforward networks, an interesting question is whether the superior performance exhibited by KAFs and 2D-KAFs can also be obtained on different architectures, such as convolutional neural networks (CNNs). To investigate this, we train several CNNs on the CIFAR-10 dataset, composed of $60000$ images of size $32 \times 32$, belonging to $10$ classes. Since our aim is only to compare different architectures for the convolutional kernels, we train simple CNNs made by stacking convolutional `modules', each of which is composed by (a) a convolutive layer with $150$ filters, with a filter size of $5 \times 5$ and a stride of $1$; (b) a max-pooling operation over $3 \times 3$ windows with stride of $2$; (c) a dropout layer with probability of $0.25$. We consider CNNs with a minimum of $2$ such modules and a maximum of $5$, where the output of the last dropout operation is flattened before applying a linear projection and a softmax operation. Our training setup is equivalent to the previous sections.

\begin{figure}
\subfloat[Test accuracy]{
\includegraphics[width=0.5\columnwidth,keepaspectratio]{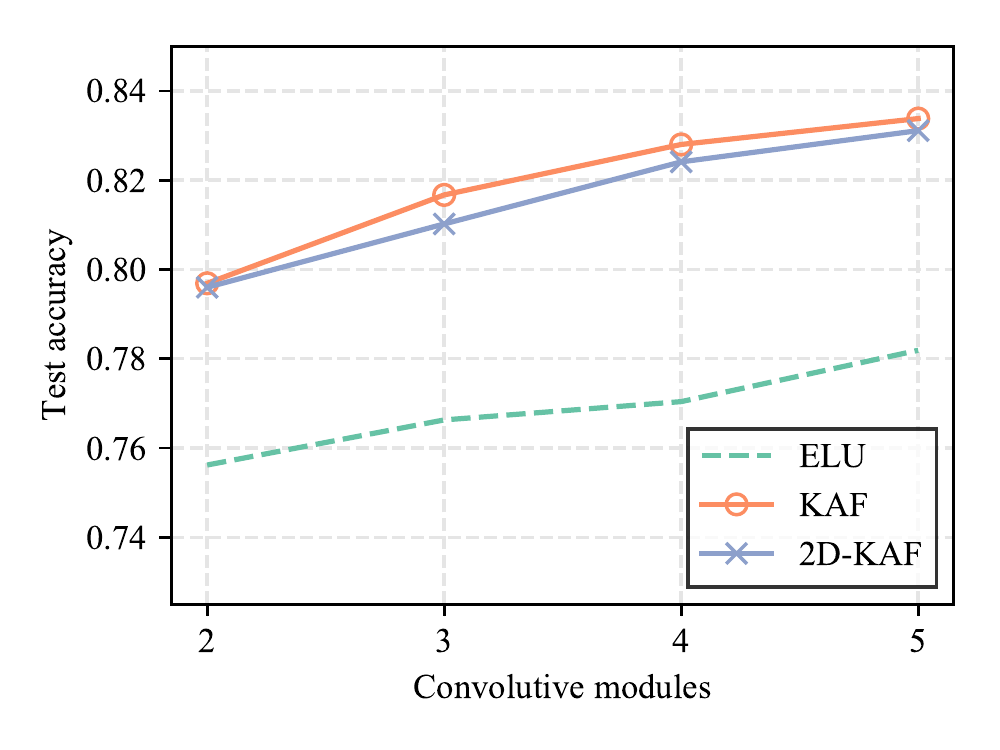}
\label{fig:cifar10_accuracy}
} \hfill
\subfloat[Parameters]{
\includegraphics[width=0.5\columnwidth,keepaspectratio]{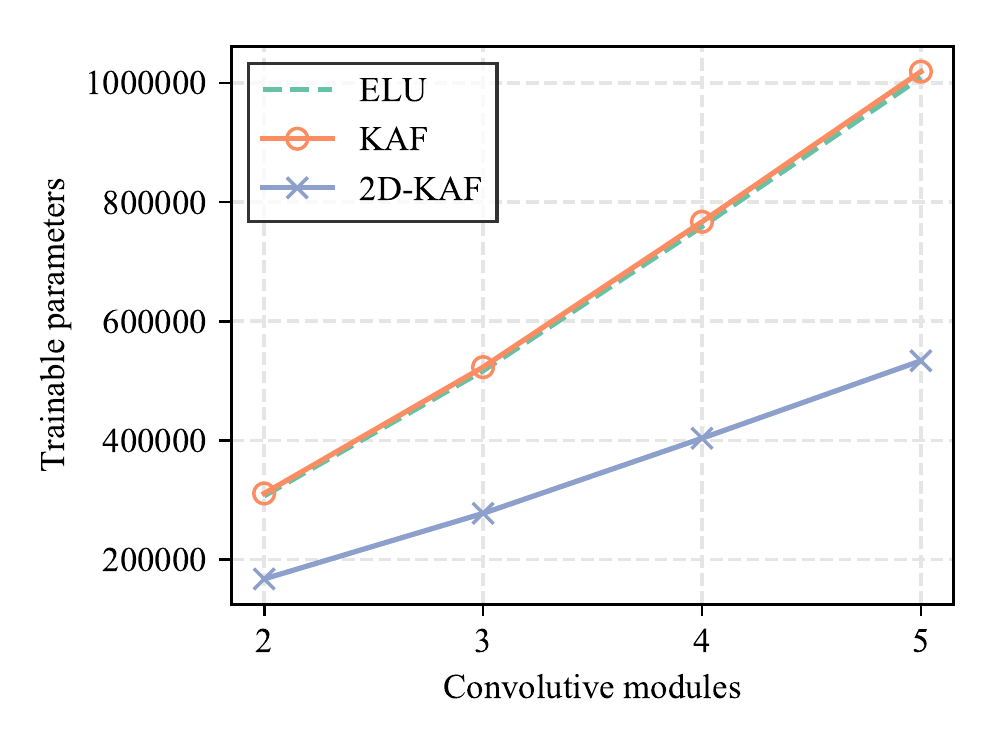}
\label{fig:cifar10_params}
} \hfill
\caption{Results of KAF, 2D-KAF, and a baseline composed of ELU functions when using only convolutive layers on the CIFAR-10 dataset (see the text for a full description of the architectures). (a) Test accuracy. (b) Number of trainable parameters for the architectures.}
\label{fig:cifar10}
\end{figure}

We consider different choices for the nonlinearity of the convolutional filters, using ELU as baseline, and our proposed KAFs and 2D-KAFs. In order to improve the gradient flow in the initial stages of training, KAFs in this case are initialized with the KRR strategy using ELU as the target. The results are shown in Fig. \ref{fig:cifar10}, where we show on the left the final test accuracy, and on the right the number of trainable parameters of the three architectures.

Interestingly, both KAFs and 2D-KAFs are able to get significantly better results than the baseline, i.e., even $2$ layers of convolutions are sufficient to surpass the accuracy obtained by an equivalent $5$-layered network with the baseline activation functions. From Fig. \ref{fig:cifar10_params}, we can see that this is obtained with a negligible increase in the number of trainable parameters for KAF, and with a significant decrease (roughly $50\%$) for 2D-KAF. The reason, as before, is that each nonlinearity for the 2D-KAF is merging information coming from two different convolutive filters, effectively halving the number of parameters required for the subsequent layer.

\subsection{Experiments on a reinforcement learning scenario}
\label{sec:results_rl}

Before concluding, we evaluate the performance of the proposed activation functions when applied to a relatively more complex reinforcement learning scenario. In particular, we consider some representative MuJoCo environments from the OpenAI Gym platform,\footnote{\url{https://github.com/openai/gym/mujoco}} where the task is to learn a policy to control highly nonlinear physical systems, including pendulums and bipedal robots. As a baseline, we use the open-source OpenAI implementation of the proximal policy optimization algorithm \citep{schulman2017proximal}, that learns a policy function by alternating between gathering new episodes of interactions with the environment, and building the policy itself by optimizing a surrogate loss function. All hyper-parameters are taken directly from the original paper, without attempting a specific fine-tuning for our algorithm. The policy function for the baseline is implemented as a NN with two hidden layers, each of which has $64$ hidden neurons with $\tanh$ nonlinearities, providing in output the mean of a Gaussian distribution that is used to select an action. For the comparison, we keep the overall setup fixed, but we replace the nonlinearities with KAF neurons, using the same initialization as the preceding sections.

\begin{figure}
\subfloat[swimmer]{
\includegraphics[width=0.3\columnwidth,keepaspectratio]{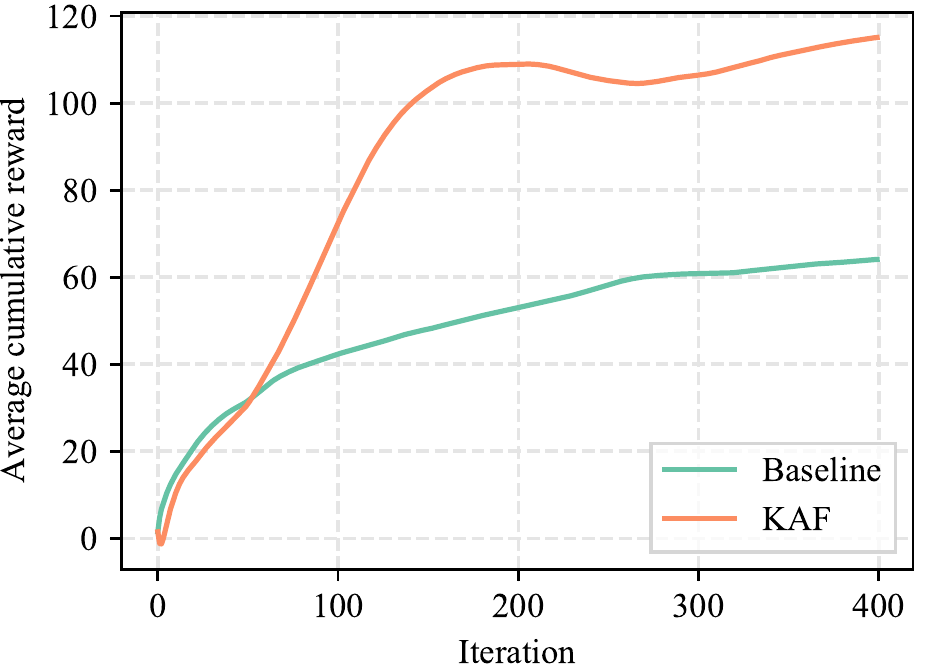}
\label{fig:reward_swimmer}
} \hfill
\subfloat[humanoidstandup]{
\includegraphics[width=0.3\columnwidth,keepaspectratio]{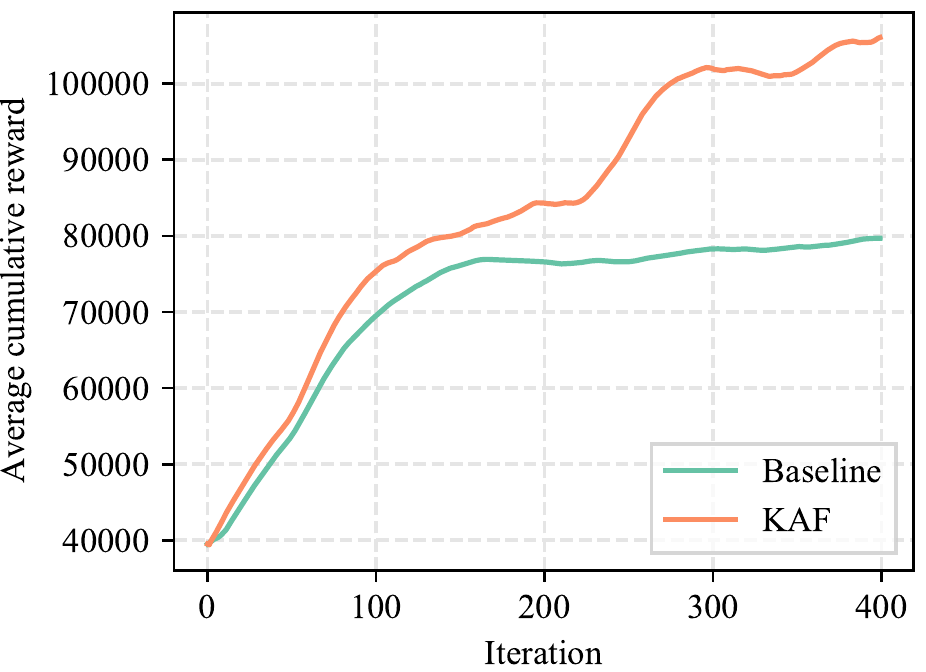}
\label{fig:reward_humanoid}
} \hfill
\subfloat[pendulum\_inverted]{
\includegraphics[width=0.3\columnwidth,keepaspectratio]{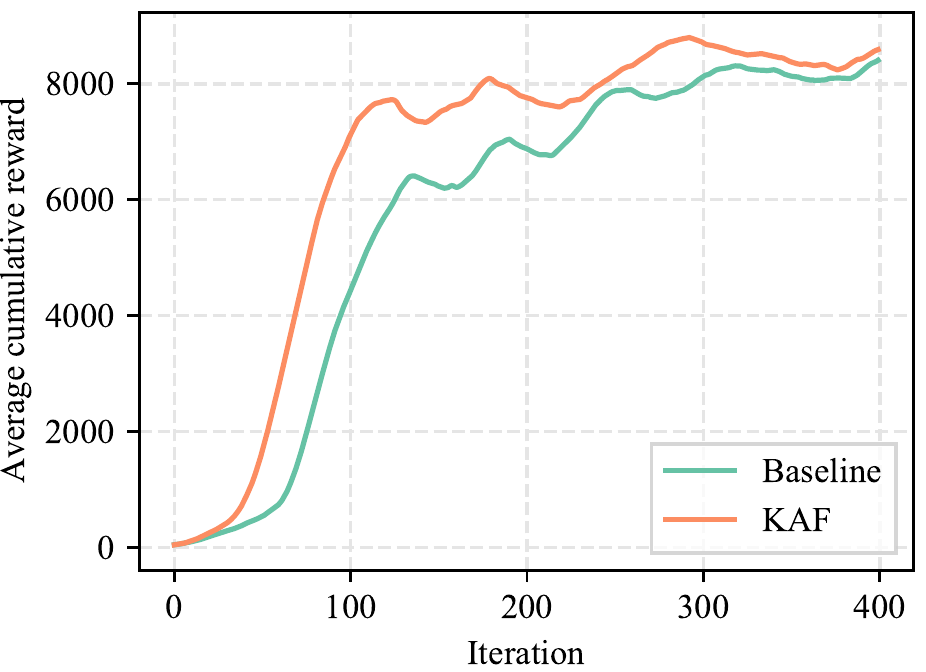}
\label{fig:reward_pendulum}
} \hfill
\caption{Results for the reinforcement learning experiments, in terms of average cumulative rewards. We compare the baseline algorithm to an equivalent architecture with KAF nonlinearities. Details on the models and hyperparameters are provided in the main discussion.}
\label{fig:rl}
\end{figure}

We plot the average cumulative reward obtained for every iteration of the algorithms on different environments in Fig. \ref{fig:rl}. We see that the policy networks implemented with the KAF functions consistently learn faster than the baseline with, in several cases, a consistent improvement with respect to the final reward.

\section{Conclusive remarks}
\label{sec:conclusions}

In this paper, after extensively reviewing known methods to adapt the activation functions in a neural network, we proposed a novel family of non-parametric functions, framed in a kernel expansion of their input value. We showed that these functions combine several advantages of previous approaches, without introducing an excessive number of additional parameters. Furthermore, they are smooth over their entire domain, and their operations can be implemented easily with a high degree of vectorization. Our experiments showed that networks trained with these activations can obtain a higher accuracy than competing approaches on a number of different benchmark and scenarios, including feedforward and convolutional neural networks.

From our initial model, we made a number of design choices in this paper, which include the use of a fixed dictionary, of the Gaussian kernel, and of a hand-picked bandwidth for the kernel. However, many alternative choices are possible, such as the use of dictionary selection strategies, alternative kernels (e.g., periodic kernels), and several others. In this respect, one intriguing aspect of the proposed activation functions is that they provide a further link between neural networks and kernel methods, opening the door to a large number of variations of the described framework.

\section*{Acknowledgments}
The authors would like to thank the anonymous reviewers for their suggestions on how to improve the paper.

\bibliographystyle{elsarticle-harv}
\bibliography{biblio}

\end{document}